\documentclass[letterpaper, 10 pt, conference]{./style/ieeeconf}
\usepackage[pdftex]{graphicx}
\graphicspath{{images-bin/}}
\usepackage[textsize=scriptsize]{todonotes}
\usepackage[nobreak]{cite}
\usepackage{svg}
\usepackage{amsmath}
\usepackage{amsfonts}
\usepackage[nohyperlinks, nolist]{acronym}
\usepackage{booktabs}
\usepackage{tabularx}
\let\labelindent\relax
\usepackage{enumitem}
\usepackage{verbatimbox}
\usepackage{csvsimple}
\usepackage{mathtools}

\usepackage{hyperref}
\usepackage{xr}
\usepackage{pifont}

\IEEEoverridecommandlockouts                             
\title{\LARGE \bf Comparing Semi-Parametric Model Learning Algorithms\\for Dynamic Model Estimation in Robotics}

\author{Sebastian Riedel$^1$, Freek Stulp$^{1,*}$
\thanks{$^1$ German Aerospace Center (DLR), Robotics and Mechatronics Center (RMC), M\"unchner Str. 20, 82234 We\ss ling, Germany, Email: \mbox{firstname.surname@dlr.de}\newline\hspace*{1em}$^*$ Corresponding author: Freek Stulp, \mbox{Freek.Stulp@dlr.de}}
}

\DeclareMathOperator*{\argmin}{arg\,min}

\begin{document}

\newcommand{\sysid}{system identification}

\newcommand{\toy}{\textsc{Toy}}
\newcommand{\toyinter}{\textsc{Toy-Interpolation}}
\newcommand{\toyextra}{\textsc{Toy-Extrapolation}}
\newcommand{\vda}{\textsc{VIA}}
\newcommand{\invda}{\textsc{VIA-Instantaneous}}
\newcommand{\arvda}{\textsc{VIA-Auto-Regressive}}
\newcommand{\simdyn}{\textsc{SimDyn}}
\newcommand{\simdynglobal}{\textsc{SimDyn-Gl}}
\newcommand{\simdynglobalinter}{\textsc{SimDyn-Gl-Interpolation}}
\newcommand{\simdynglobalextra}{\textsc{SimDyn-Gl-Extrapolation}}
\newcommand{\simdynlocal}{\textsc{SimDyn-Ll}}
\newcommand{\simdynlocalinter}{\textsc{SimDyn-Ll-Interpolation}}
\newcommand{\simdynlocalextra}{\textsc{SimDyn-Ll-Extrapolation}}

\newcommand{\mytodo}[1]{\textbf{\textcolor{red}{TODO: #1}}}

\newcommand{\spgp}{\textsc{SPGP}}
\newcommand{\gp}{\textsc{GP}}
\newcommand{\bambann}{\textsc{BaMbANN}}
\newcommand{\bnn}{\textsc{BNN}}
\newcommand{\lls}{\textsc{LLS}}
\newcommand{\llsgp}{\textsc{LLS-GP}}
\newcommand{\llsbnn}{\textsc{LLS-BNN}}
\newcommand{\itllsgp}{it-\textsc{LLS-GP}}
\newcommand{\itllsbnn}{it-\textsc{LLS-BNN}}
\newcommand{\svr}{\textsc{SVR}}
\newcommand{\svrgp}{\textsc{SVR-GP}}
\newcommand{\svrbnn}{\textsc{SVR-BNN}}
\newcommand{\itsvrgp}{it-\textsc{SVR-GP}}
\newcommand{\itsvrbnn}{it-\textsc{SVR-BNN}}

\begin{acronym}[SPML]
\acro{RL}{reinforcement learning}
\acro{GP}{Gaussian process}
\acro{NN}{neural network}
\end{acronym}

\maketitle
\setcounter{footnote}{1}

\begin{abstract}
Physical modeling of robotic system behavior is the foundation for controlling many robotic mechanisms to a satisfactory degree.
Mechanisms are also typically designed in a way that good model accuracy can be achieved with relatively simple models and model identification strategies.
If the modeling accuracy using physically based models is not enough or too complex, model-free methods based on machine learning techniques can help.
Of particular interest to us was therefore the question to what degree semi-parametric modeling techniques, meaning combinations of physical models with machine learning, increase the modeling accuracy of inverse dynamics models which are typically used in robot control.
To this end, we evaluated semi-parametric Gaussian process regression and a novel model-based neural network architecture, and compared their modeling accuracy to a series of naive semi-parametric, parametric-only and non-parametric-only regression methods.
The comparison has been carried out on three test scenarios, one involving a real test-bed and two involving simulated scenarios, with the most complex scenario targeting the modeling a simulated robot's inverse dynamics model.
We found that in all but one case, semi-parametric Gaussian process regression yields the most accurate models, also with little tuning required for the training procedure.
\end{abstract}

\section{Introduction}

Robot control benefits greatly from an accurate dynamics model, so that known forces can be compensated for through feedforward control.
Good feedforward control means feedback control must only compensate for \emph{unknown} perturbations and as a result, for example, low-gain compliant control becomes feasible.

Since humans design robots\footnote{Unless their morphology is evolved~\cite{lipson2000automatic}}, models used for the design and construction alongside physical modeling of the robot (rigid body dynamics, actuator dynamics, etc.) can be used to determine the structure and parameters of dynamic models.
Thus, to us, it seems superfluous to learn dynamic models from scratch when good models are already available. However, good does not mean perfect. 

Analytical models are abstraction, and include assumptions that do not hold in the real world (link inertias are known exactly, cables do not need to be taken into account, friction is not temperature-dependent).
The models also cannot account for local variations e.g. due to small manufacturing flaws or wear and tear of the robot.
Accurately estimating the parameters of such a model from data, which we refer to as \emph{parametric model identification}, is a tedious and non-trivial task on its own the more complex a model gets.
For example, see the comments and results of the learning-based approach of \cite{hwangbo2019learning} with respect to their previous (complex) model-based approach \cite{gehring2016practice} in controlling a quadruped robot.

Purely non-parametric approaches to obtaining a dynamics model for robot control lie on the other end of the spectrum \cite{nguyen-tuong2008computed}.
While such approaches often lead to higher accuracy on specific trajectories, generalization to unseen parts of the configuration space (with respect to the training data distribution) is a challenge. 
We use the term \emph{non-parametric model learning} to refer to approaches of this kind.

\begin{figure}[t]
  \centering
  \includegraphics[width=.4\textwidth]{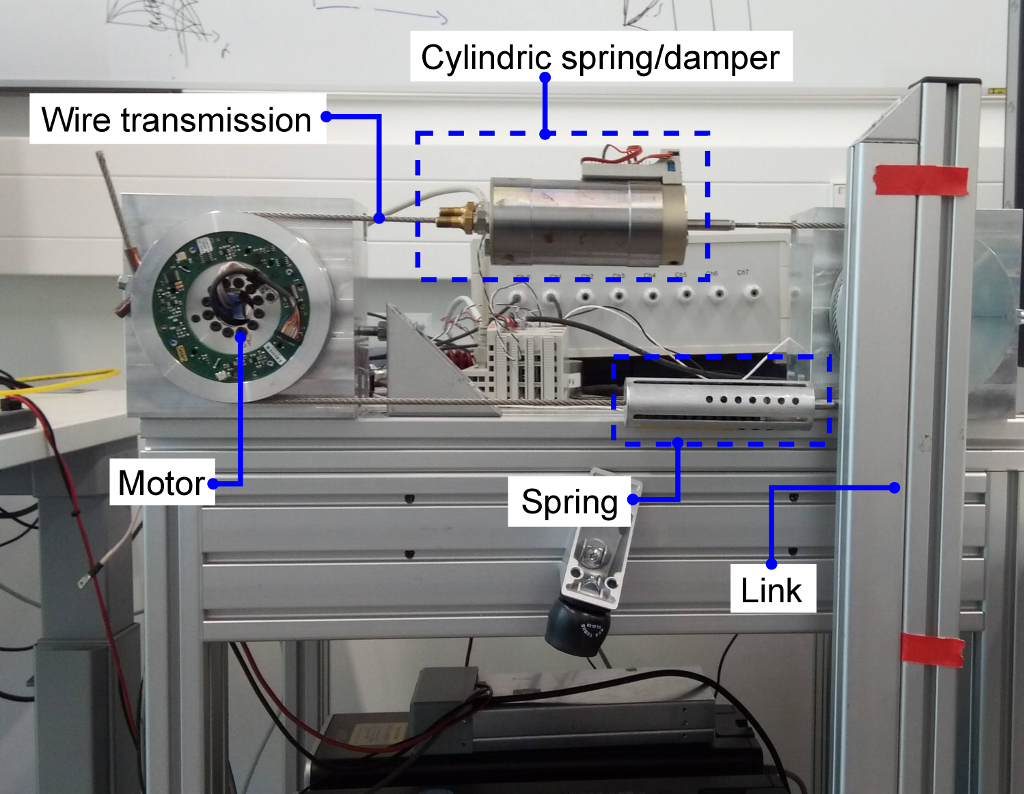}
  \caption{Physical setup for the variable impedance actuator test scenario (\vda{}) described in section \ref{sec:scenariosvia}}
  \label{fig:viatestbed}
\end{figure}

The strengths and weaknesses of both approaches are summarized in Table~\ref{tab:strenghts}. We think, the best of both worlds can be achieved by combining global parametric models, in which we can include our knowledge of the robot and physics, with non-parametric models, which can account for local deviations from the parametric model. The aim of this paper is therefore to extensively compare different variants of such \emph{semi-}parametric approaches to model learning.

Our main contributions with respect to already published articles about non- or semi-parametric model learning, e.g. \cite{nguyen-tuong2008computed, nguyen-tuong2010using, wu2012semi, krishnamoorthi2018model, camoriano2016incremental}, are three-fold:
\begin{itemize}
\item we evaluate well-known semi-parametric approaches (like semi-parametric Gaussian processes) to a wider variety of baselines, including e.g. simpler semi-parametric approaches as well as robust model identification through model-based support vector regression
\item we include evaluation of a semi-parametric Bayesian neural network approach
\item all experiment data (split into the training and testing sets we used) as well as the parametric models for each evaluation scenario are published along-side this letter for own experiments or reproducing our results \footnote{\url{https://rmc.dlr.de/download/semiparametric-learning/spml_data.zip}}
\end{itemize}

In the next section, we give a more formal introduction to the problem we study. In Section \ref{sec:methods}, the evaluated methods are described in detail and in Section \ref{sec:scenarios} so are our evaluation scenarios. Section \ref{sec:results} discusses our results and observations. In \ref{sec:rw}, related work in the area of (semi-)parametric model learning and successful applications thereof are put into context with our evaluated methods here, and lastly, we conclude our findings. 

\section{Problem Formulation}
\label{sec:problem}

This letter summarizes results concerning model learning in settings for which an analytical parametric model exists.
The process of model learning then includes i) calibrating the analytical model's parameters to some data (\emph{parametric model identification}, e.g. using linear least squares) and ii) learning residual errors between observations and the (calibrated) analytical model via a non-parametric machine learning approach (e.g. Gaussian processes).
We term learning approaches which deal with i) and ii) together as \emph{semi-parametric model learning}, where \emph{parametric} refers to the parameters of the analytical model and \emph{semi} to the additional non-parametric machine learning component. Further, we distinguish between \emph{joint} semi-parametric and \emph{sequential} semi-parametric approaches depending on if both aspects are learned in a joint objective or if they are dealt with in a sequential, potentially iterative way.

Formally, learning a model in our context refers to estimating a function $y = f(x), x \in \mathbb{R}^{D_{in}}, y \in \mathbb{R}^{D_{out}}$, given observations $X = \{x_0, \ldots, x_N\}, Y = \{y_0, \ldots, y_N\}$ and some prior analytical model $y = h_{\theta_m}(x)$ with $\theta_m \in \mathbb{R}^M$ being the to-be-calibrated model parameters.

In semi-parametric model learning, we define

\begin{equation}
f(x) \approx f_{\{\theta_m,\theta_{np}\}} = h_{\theta_m}(x) + g_{\theta_{np}}(x)
\label{eq:spml}
\end{equation}

which expresses the desire that the non-parametric model $g_{\theta_{np}}(x)$ with parameters $\theta_{np}$ is responsible for representing the residual error between the observed data and the parametric model $h_{\theta_m}(x)$.

The model $h_{\theta_m}(x)$ in all our evaluation scenarios has a linear-in-parameters form for each output dimension $y_i, i \in [1,\ldots,D_{out}]$

\begin{align}
y_i &= \phi_i(x)^T\theta_m\\
\phi_i(x)&: \mathbb{R}^{D_{in}} \rightarrow \mathbb{R}^M
\end{align}

While this linear-in-parameters form plays nicely with the often used regressor formulation for parametric model identification in the context of rigid body dynamics~\cite{khalil1997symoro}

\begin{equation}
\tau = \Phi(q, \dot{q}, \ddot{q})\theta
\end{equation}

it is not necessary for our presented joint semi-parametric approaches, which could deal with non-linear-in-parameters models as well, as long as the model is differentiable with respect to the parameter vector $\theta$.

\begin{table*}[t]
\begin{tabularx}{\textwidth}{X|X|X}
\toprule
\multicolumn{1}{c|}{\textbf{Approach}} & \multicolumn{1}{c|}{\textbf{Strength}} & \multicolumn{1}{c}{\textbf{Weakness}} \\ \midrule
parametric model identification
&
\vspace{-\baselineskip}\begin{itemize}[leftmargin=8pt]
\item few training data required
\item interpretability of estimated model parameters
\item usually very fast (identification and evaluation)
\item can be designed to meet specific requirements, e.g. smoothness or analytic derivatives
\item good at extrapolation outside training data (within the parametric models validity)
\end{itemize}
&
\vspace{-\baselineskip}\begin{itemize}[leftmargin=8pt]
\item representational power limited to parametric model's structure
\item significant engineering effort to build accurate models
\item model identification from data requires excitation (and observability) of all parameters
\end{itemize} \\ \midrule
non-parametric model learning
&
\vspace{-\baselineskip}\begin{itemize}[leftmargin=8pt]
\item arbitrarily accurate as representational power can be scaled with available amount of training data
\end{itemize}
&
\vspace{-\baselineskip}\begin{itemize}[leftmargin=8pt]
\item usually much more training data required
\item bad at extrapolating outside training data
\end{itemize} \\ \bottomrule
\end{tabularx}
\caption{Relative strengths and weaknesses for parametric and non-parametric model learning.}
\label{tab:strenghts}
\end{table*}

\section{Methods}
\label{sec:methods}

Of particular interest to us was the comparison of joint vs. sequential semi-parametric model learning.
Following equation \eqref{eq:spml}, we categorize an approach as joint semi-parametric if some loss objective $\mathcal{L}$ for given observed data $\{X, Y\}$ is minimized jointly over both parameter sets $\{\theta_m, \theta_{np}\}$

\begin{equation}
\{\theta_m^*,\theta_{np}^*\} = \argmin_{\theta_m,\theta_{np}}\mathcal{L}(f_{\{\theta_m,\theta_{np}\}}(X), Y)
\end{equation}

Sequential semi-parametric approaches are defined as two-step process.
First, the parametric model is fit to observed data and the target residuals $Y^*$ with respect to the estimated parametric model are computed \eqref{eq:pfit}-\eqref{eq:pfitresidual}.
Second, the non-parametric model is fit to the residuals \eqref{eq:npfit}.
For predictions, both estimated models are then combined as in equation \eqref{eq:spml}.

\begin{align}
\theta_m^* &= \argmin_{\theta_m}\mathcal{L}(h_{\theta_m}(X), Y) \label{eq:pfit}\\
Y^* &= \{y^*_i \mathrm{~with~} y^*_i = y_i - h_{\theta_m^*}(x_i), i \in [0, N]\} \label{eq:pfitresidual}\\
\theta_{np}^* &= \argmin_{\theta_{np}}\mathcal{L}'(g_{\theta_{np}}(X), Y^*) \label{eq:npfit}
\end{align}

We would expect, that in general optimizing for both parameter sets jointly would outperform sequential learning approaches.
Furthermore, iterative sequential semi-parametric approaches have been implemented by iterating equations \eqref{eq:pfit}-\eqref{eq:npfit} a fixed number of times and fitting the parametric model in equation \eqref{eq:pfit} to the residual error  $Y'$ of the data with respect to the current iteration's non-parametric model.

\begin{equation}
Y' = \{y'_i \mathrm{~with~} y'_i = y_i - g_{\theta_{np}^*}(x_i), i \in [0, N]\} \label{eq:npfitresidual}
\end{equation}

\subsection{Semi-Parametric Gaussian Processes (\spgp{})}
The main algorithm we wanted to evaluate and put into perspective to other approaches is semi-parametric Gaussian process regression.

Gaussian processes offer a convenient and principled way to incorporate a parametric model into the learning process \cite[chapter~2.7]{rasmussen2003gaussian}.
As a \ac{GP} is described by a mean $\mathit{m}(x)$ and kernel function $\mathit{k}_{\theta_k}(x, x')$, it suffices to replace $\mathit{m}(x)$ with the parametric model in order to obtain a semi-parametric model which follows the formulation in equation \eqref{eq:spml} (for the predictive mean of the \ac{GP}).
If our model (resp. mean function) has open parameters $\theta_m$, notated as $\mathit{m}_{\theta_m}(x)$, and depending on if the model is linear or non-linear in those parameters, we have different choices for inferring the parametric model's parameters as well as the kernel hyper-parameters from data.

The first approach works for linear- as well as non-linear-in-parameters mean functions and jointly minimizes the log marginal likelihood of the \ac{GP} model with respect to the total parameter set $\theta = \{\theta_m, \theta_k\}$.
If the mean function is differentiable, as in all our cases, this can be done efficiently using a gradient-based optimizer.
We term this variant Semi-Parametric GP (\spgp{}) with joint optimization.

For linear-in-parameters parametric models (mean functions) and assuming a Gaussian prior $\theta_m \sim \mathcal{N}(b, B)$ on the model parameters, \cite{rasmussen2003gaussian} (based on \cite{ohagan1978curve}) shows that it is possible to marginalize out the model parameters analytically.
This could be especially interesting in the limiting case of a very vague prior on those parameters (infinite variance $B$).
We implemented this variant as well, but as preliminary experiments showed little difference between the joint-optimization variant and the marginalized-out variant for the (much lower than infinity) prior uncertainty we expected on model parameters, we only report results for the more general joint-optimization variant.
Also, \cite{rasmussen2003gaussian} points out that the for the limiting case of infinite variance additional care regarding the numerical side of the implementation has to be taken and we did not want to deal with that problem for now.

We have implemented our \spgp{} with joint $\{\theta_m, \theta_k\}$-parameter optimization using the library GPflow \cite{matthews2017gpflow} as well as using the C++ library Limbo \cite{cully2018limbo}. If not noted otherwise, results were obtained using the implementation in GPflow.
For the \spgp{} (as well as baseline \gp{}s), a standard RBF-kernel with automatic relevance determination and a Gaussian observation likelihood were used.
RBF kernel hyper-parameters (length scales and variance) as well as the observation likelihood variance were initialized to 1.0 and optimized using GPflow's default optimizer (L-BFGS-B).
Initialization values for the mean function parameters $\theta_m$ depend on the test scenario.

\subsection{Bayesian Model-based Neural Networks (\bambann{})}
In an effort to tackle some of the problems of Gaussian Process regression, e.g. undesired computational scaling with available data, smoothness assumptions when using standard stationary kernels like RBF and Matern, and the usually naively independent modeling of systems which require vector predictions (not scalar ones), we tried to blend neural networks with parametric models.

Neural networks (NNs) are well known as universal function approximators defined by a graph of typically layer-wise computations whose parameters are optimized using variants of stochastic gradient descent with respect to some training loss.
In this letter, we solely use dense feedforward NNs.
We incorporate the parametric model into the NN by calculating it in a parallel sub-graph of the network and summing the regular dense path with the parametric prediction to form the final output.
Following \eqref{eq:spml}, the combined output of the network is therefore

\begin{equation}
y = f_{model,\theta_m}(x) + f_{dense,\theta_{dense}}(x)
\end{equation}

With standard stochastic gradient descent and a supervised loss (e.g. mean squared error), we can adapt the NN parameters $\theta_{dense}$ jointly with calibrating the model parameters $\theta_m$.
However, trained in this way, the combined model has no incentive to use the dense layers solely for modeling the residual between the parametric model and the data (one could play with different regularization strengths for model vs. NN parameters, but we did not try this).
Therefore we adopt the Bayesian neural network (\bnn{}) formulation described in \cite{blundell2015weight}, where a probability distribution over the network's parameters is optimized instead of a point-estimate. The loss in this formulation consists of data likelihood term, driving the model's weights to fit the data more closely, as well as a KL-term keeping the weight distributions close to a given prior.
We use the prior distributions to regularize the network in its use of the dense NN layers by using a small, zero-centered Gaussian prior for the dense weights $\theta_{dense}$ and a wide, (potentially non-zero centered) Gaussian prior for the model weights $\theta_{m}$.
As the trained \bnn{} represents the dense and parametric model parameters by a probability distribution, we obtain predictions for an input by sampling all network parameters 30 times and calculating the sample mean and variance over the network's predictions for the 30 parameter sets.

We denote the resulting model as Bayesian Model-based artificial neural network (\bambann{}) and implemented it on top of the \bnn{} layers provided by tensor2tensor \cite{vaswani2018tensor2tensor}.

In general, we used ADAM (with keras' default parameters) as optimizer and a 3-layer feedforward NN architecture with \emph{elu} non-linearities (output layer with \emph{linear} activation). Each hidden layer had 64 (\toy{} \& \vda{} scenarios) or 128 neurons (\simdyn{} scenario).
For the data likelihood term, we used a Gaussian observation likelihood with a fixed observation variance.
We found that this observation variance has a large impact on the learning performance and generally needed to be set to a much lower value than the ``real'' observation noise in order for the \bnn{}s to start fitting the data at all (something which warrants more investigation in the future).
For the dense path through the \bambann{} network, standard scaling\footnote{scikit-learn's \emph{sklearn.preprocessing.StandardScaler}} has been applied.

\subsection{Baselines: Solely Parametric and Non-Parametric Learning, and Combinations}

We divide our baselines into three categories: i) solely parametric, ii) solely non-parametric and iii) sequential semi-parametric (in contrast to the joint semi-parametric methods \spgp{} and \bambann{}).
\subsubsection{Parametric Baselines}
As solely parametric baselines, we use the analytic model alone and fit its parameters to data either via linear least squares (\lls{}) or support vector regression \cite{smola2004tutorial} (\svr{}).
This works, because for all three test scenarios the parametric model is in fact linear-in-parameters.
For \svr{}, we define a kernel using the model as basis function (only the model, nothing more) thus the main difference between \lls{} and \svr{} is that \lls{} minimizes a squared loss and \svr{} a linear one, making it more robust to outliers.
\lls{} is implemented using plain Numpy/Python, for \svr{} we use scikit-learn's implementation.

\subsubsection{Non-Parametric Baselines}
As solely non-parametric baselines, we use a \gp{} and a dense feed-forward \bnn{} in their respective plain versions without any parametric model.

\subsubsection{Sequential Semi-Parametric Baselines}
Sequential semi-parametric approaches are obtained by combining the previously mentioned baseline methods.
By first fitting either a solely parametric \lls{} or \svr{} regressor and then fitting the residual error using either a solely non-parametric \gp{} or \bnn{}, baseline variants denoted as \llsgp{}, \llsbnn{}, \svrgp{} and \svrbnn{} are formed.
As described at the beginning of \ref{sec:methods}, we also tried naively iterating these two fitting steps. These approaches are denoted by the prefix \emph{it-}, e.g. \itllsgp{}.

\section{Experiment Scenarios}
\label{sec:scenarios}

\subsection{1d Toy Example (\toy{})}
The toy example consists of a 1d regression problem ($D_{in} \coloneqq 1, D_{out} \coloneqq 1, M \coloneqq 4$) with noisy samples from a linear-in-parameters model $\phi(x)^T\theta_m$ with a strong (but local and smooth) deviation $f_{dev}(x)$ of the data from the model around $x \coloneqq 2.5$.

\begin{equation}
y_{obs} = \phi(x)^T\theta_m + f_{dev}(x) + \epsilon
\end{equation}

with

\begin{align}
\phi(x)^T &= \begin{pmatrix}sin(2x)\\x\\1.0\\0.09x^2\end{pmatrix}^T\\
\theta_m &\in \mathbb{R}^4\\
\epsilon &= \mathcal{N}(\mu=0,\sigma=0.5)
\end{align}

\begin{figure}[htbp]
  \centering
  \includegraphics[width=.4\textwidth]{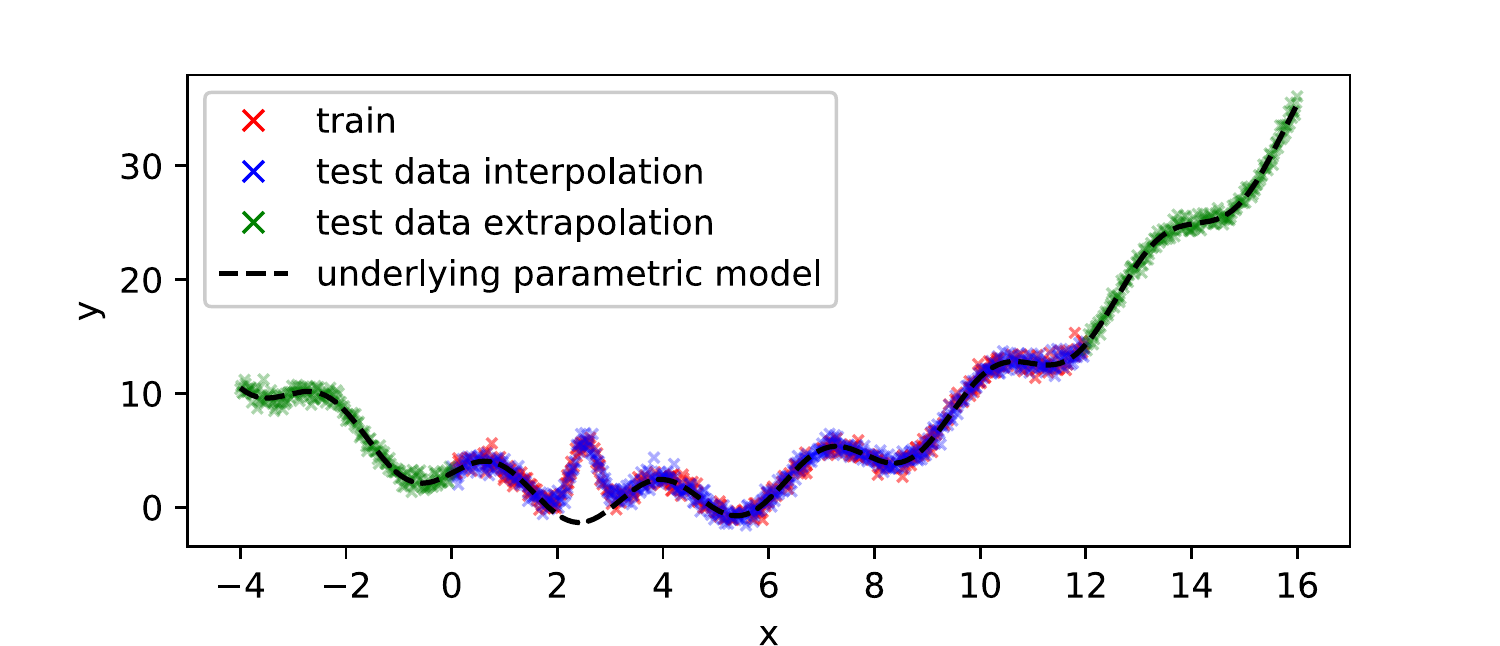}
  \caption{Training data and the two test datasets \toyinter{} and \toyextra{}.}
  \label{fig:toydataset}
\end{figure}

The model coefficients are set to $\theta_m = (2.0, -1.5, 3.0, 2.4)^T$ and training data is derived by sampling the function $N=400$ times uniformly between $x \in [0.0, 12.0]$.
An \textit{interpolation} test data set is created by independently sampling the function again in the interval $x \in [0.0, 12.0]$ giving rise to the test scenario \toyinter{}.
Additionally, a second set of test data is created by sampling the function in the intervals $x \in \{[-4.0, 0.0],[12.0, 16.0]\}$ in order to assess the \textit{extrapolation} performance of the different approaches. This scenario is denoted as \toyextra{}. Figure \ref{fig:toydataset} illustrates the training and two test datasets.

For \spgp{} and \bambann{}, optimization started from zero-valued parametric coefficients $\theta_m = \boldsymbol{0} \in \mathbb{R}^4$. \lls{} and \svr{} do not require initial values.

\subsection{Variable Impedance Actuator (\invda{}, \arvda{})}
\label{sec:scenariosvia}
This scenario is based on data collected from a variable impedance actuator (VIA) test-bed.
The physical setup consists of one VIA joint, meaning one motor connected to one link via a parallel arrangement of a fixed-stiffness spring and a variable-damping damper.
An illustration of the mechanism is given in Figure \ref{fig:viatestbed} and more details can be found in \cite{kim2017enhancing, garofalo2019joint}.
In modeling this system, the link-side torque $\tau$ equals the torque produced by the sum of the spring and damping mechanisms as in 

\begin{equation}
\tau = D(t)(\dot{\theta} - \dot{q}) + K(\theta - q)
\end{equation}

where the difference in motor and link position ($\theta - q$) and velocity ($\dot{\theta} - \dot{q}$) result in the respective spring and damping torques through the stiffness coefficient $K$ and the - possibly time-varying - damping coefficient $D(t)$.
During data collection for the experiment here, the damping was set a fixed value (time-invariant $D$).
The regression problem is then formulated with $\tau$ as target, the position and velocity difference as input and $\{K, D\}$ (time-invariant) as the two parametric model coefficients, resulting in the dimensionalities $D_{in}\coloneqq 2, D_{out} \coloneqq 1, M \coloneqq 2$ and denoted as \invda{}.

As we found that the system exhibits considerable stick-slip friction, a phenomenon which cannot be modelled using only instantaneous motor and joint measurements at a single time-step, we set up a second, auto-regressive regression problem where the input is augmented by the measurements of the last four time-steps in addition to the current one. This regression setup improved prediction quality considerably and we denote this problem as \arvda{} where the dimensionalities are $D_{in} \coloneqq 10, D_{out} \coloneqq 1, M \coloneqq 2$.

The data was collected by letting a motor position controller follow a chirp-signal (sinusoidal trajectory with increasing frequency) from zero to three~Hz. Telemetry was recorded at 1~kHz, resampled to 100~Hz and split at round 64\% through the chirp motion to form the training (first 64\% percent) and test data (remaining 36\%). The history of samples for the \arvda{} case thus reaches back $40~\mathrm{ms}$ at a $10~\mathrm{ms}$ interval.

For \spgp{} and \bambann{}, optimization started from $(K, D) = (300.0, 20.0)$, if not noted otherwise, and the parameters were expected to be around $(K, D) \approx (400.0, 10.0)$. \lls{} and \svr{} do not require initial values.

\subsection{Simulated Robot Dynamics (\simdyn{})}
This scenario is about learning an inverse dynamics model for a simulated, planar three-link robot. The inverse dynamics function of a robot is essential for precise control over the robot's motion and relates desired joint accelerations $\ddot{q}$ at a given joint position and velocity state $\{q, \dot{q}\}$ to the necessary joint torques $\tau$ which need to be applied to achieve that acceleration.
The underlying regression problem here has the dimensionalities $D_{in} \coloneqq 9, D_{out} \coloneqq 3$ with $M \coloneqq 17$ parametric model coefficients.

To test semi-parametric learning methods in this context, we developed a simulation environment around the robot modeling toolbox OpenSyMoRo \cite{khalil2014opensymoro}.
In particular, OpenSyMoRo is used to calculate the analytic expressions for the forward dynamics model, which are then used in the simulation of the system.
It also calculates the expressions for the dynamic identification regressor $\Phi$ together with a vector of model coefficients $\theta_m$, which serve as the parametric model for all (semi-)parametric methods described here.
For a general N-link serial structure robot following the classic rigid body dynamics equations, the inverse dynamics function can be written in a linear-in-parameters regressor form $\tau = \Phi(q, \dot{q}, \ddot{q})\theta_m$, where $\theta_m$ is the $N*13$-dimensional parameter vector of this model.
These 13 parameters per joint specify the physical dynamical parameters of the system and consist of link inertial properties, link mass, motor inertia, Coulomb and viscous joint friction as detailed in \cite{khalil1997symoro}.

In the case of the planar three-link robot, the dimensionality of the involved quantities is as follows. $\theta_m \in \mathbb{R}^{17}$ holds the 17 model parameters (in re-grouped form, because not all $3*13=39$ parameters can be estimated independently or are even observable/relevant for a planar, 100\% rigid robot), $q, \dot{q}, \ddot{q} \in \mathbb{R}^3$ represent the joint positions, velocities and accelerations, and $\tau \in \mathbb{R}^3$ the matching joint torques for all three joints.

A simulated robot in our setting is then defined by a configuration $R = (\theta_m, f_{mis})$ with parametric model coefficients $\theta_m$ and a configured model-mismatch $f_{mis}$ (additional dynamics which are not part of the model structure $\Phi$). 

For $f_{mis}$, we implemented two variants.
In the first setting, similar to the \toy{} scenario, a local (in input-space) model-mismatch was introduced by adding a strong friction torque to the first joint between $0.15~\mathrm{rad} \leq q_0 \leq 0.25~\mathrm{rad}$.
We denote variants using this mismatch model as \simdynlocal{} (\textbf{L}oca\textbf{l}).
As only the first joint is affected, the regression problem for the other joints should not require any non-parametric learning at all (cf. no model mismatch).

In a second variant, in addition to the local friction on joint zero, we add a global, joint-value dependent ripple torque $e_i$ to each joint $i \in [0,1,2]$ given by

\begin{equation}
e_i = a_1 \sin(h_1 M q_i) + a_2 \sin(h_2 M q_i)
\end{equation}

This loosely mimics the model of harmonic drive ripple torque described in \cite{godler1999ripple} for two frequency multiples $h_1=2, h_2=8$ with amplitudes $a_1=6.0, a_2=2.0$ and a virtual gear reduction of $M=30$.
While for the first joint the model error along the tested trajectories is dominated by the local friction, the model errors on the other joints are produced due to the ripple torque and, as we found, are a challenge for all tested modelling/learning methods.
Data obtained using this mismatch model is denoted by the prefix \simdynglobal{} (\textbf{G}loba\textbf{l}).

The general procedure to obtain training and test data is then given by combining a simulated robot $R = (\theta_m, f_{mis})$ (model coefficients with mismatch setting) with a model-based inverse dynamics controller $C = \big((q_{des}, \dot{q}_{des}, \ddot{q}_{des}), q_{\mathrm{offset},0}, \hat{R}, (k_p, k_d)\big)$ which tries to follow an oscillatory excitation motion $(q_{des}, \dot{q}_{des}, \ddot{q}_{des})$ around a start offset $q_{\mathrm{offset},0}$ (on the first joint only) using an internal robot model $\hat{R}$ and controller gains $(k_p, k_d)$.

We simulated four trajectory rollouts by combination of $q_{\mathrm{offset},0} \in \{0.0\mathrm{rad}, 0.3\mathrm{rad}\}$ and $f_{mis} \in \{\mathrm{local}, \mathrm{global}\}$. The different settings for $f_{mis}$ represent two separate settings in which we want to evaluate all methods. The different reference positions $q_{\mathrm{offset},0}$ around which the trajectory will oscillate are used to provide inter- and extrapolation test datasets.
Each roll-out is 100 seconds with control and sample rate of 1~kHz.
The first 50 seconds of each mismatch variant and $q_{\mathrm{offset},0} = 0.0\mathrm{rad}$ are subsampled to 100~Hz and used as training data (subsampling ensures a modest training data size of $\approx$ 5000 samples which can still be handled well by \gp{}-based methods).
The last 50 seconds in the same $q_{\mathrm{offset},0} = 0.0\mathrm{rad}$-setting are used as interpolation test data and form the test datasets \simdynlocalinter{} and \simdynglobalinter{}.

The last 50 seconds of each $q_{\mathrm{offset},0} = 0.3\mathrm{rad}$-setting are used as extrapolation test data and form the test datasets \simdynlocalextra{} and \simdynglobalextra{}.

In general, the resulting motion $q, \dot{q}, \ddot{q}$ as well as the commanded torque $\tau$ were recorded to form the datasets, where $\dot{q}, \ddot{q}$ are obtained by numeric differentiation with a modelled joint position sensor noise on $q$ uniformly distributed between $\pm 1e^{-6}~\mathrm{rad}$.

For the excitation motion design, we follow \cite{nguyen-tuong2008computed} where each desired joint position is the sum of two sinusoidals $q_i(t) = A_i sin(2 \pi f_{1i} t) + A_i/3 sin(2 \pi f_{2i} t)$ with $A_i = [0.4, 0.2, 0.3], f_{1i}=[0.28, 0.52, 0.26], f_{2i}=[1.1, 2.3, 2.2]$.
The controller implements a standard inverse dynamics control approach in a high-gain setting using $k_p=3947.8, k_d=125.7$.

Table \ref{tab:simdyncoeffs} shows the ground-truth coefficients as well as the initial values from which the optimization started for \spgp{} and \bambann{}, if not noted otherwise. \lls{} and \svr{} do not require initial values.

\begin{table}
\centering
\caption {\label{tab:simdyncoeffs} Analytic model coefficients used in the simulated model (ground truth) vs. coefficients used in the controller (prior). Prior values are also used as optimization starting point for model learning.}
\begin{tabular}{l||lll}
Coefficient & Truth & Prior & $|\mathrm{Deviation} [\%]|$\\
\hline
\begin{filecontents*}{./csvs/simdyn_model_coeffs.csv}
prior,groundtruth,names,deviationperc
0.00,0.00,FS1,0.00
0.00,0.00,FS2,0.00
0.00,0.00,FS3,0.00
10.17,10.00,FV1,1.74
10.59,10.00,FV2,5.89
10.07,10.00,FV3,0.72
1.02,1.00,IA2,1.94
1.06,1.00,IA3,5.81
0.11,0.10,MX3,6.38
9.35,8.83,MXR1,5.86
3.34,3.83,MXR2,12.79
0.00,0.00,MY1,0.00
0.00,0.00,MY2,0.00
0.00,0.00,MY3,0.00
0.12,0.10,ZZ3,16.29
10.01,9.43,ZZR1,6.10
2.92,3.43,ZZR2,14.92
\end{filecontents*}
\csvreader[head to column names]{./csvs/simdyn_model_coeffs.csv}{}
{\\\names & \groundtruth & \prior & \deviationperc ~$\%$}
\end{tabular}
\end{table}

\section{Results and Discussion}
\label{sec:results}

\subsection{\toy{} Scenario}

\begin{figure}[htbp]
  \centering
  \includegraphics[width=.5\textwidth]{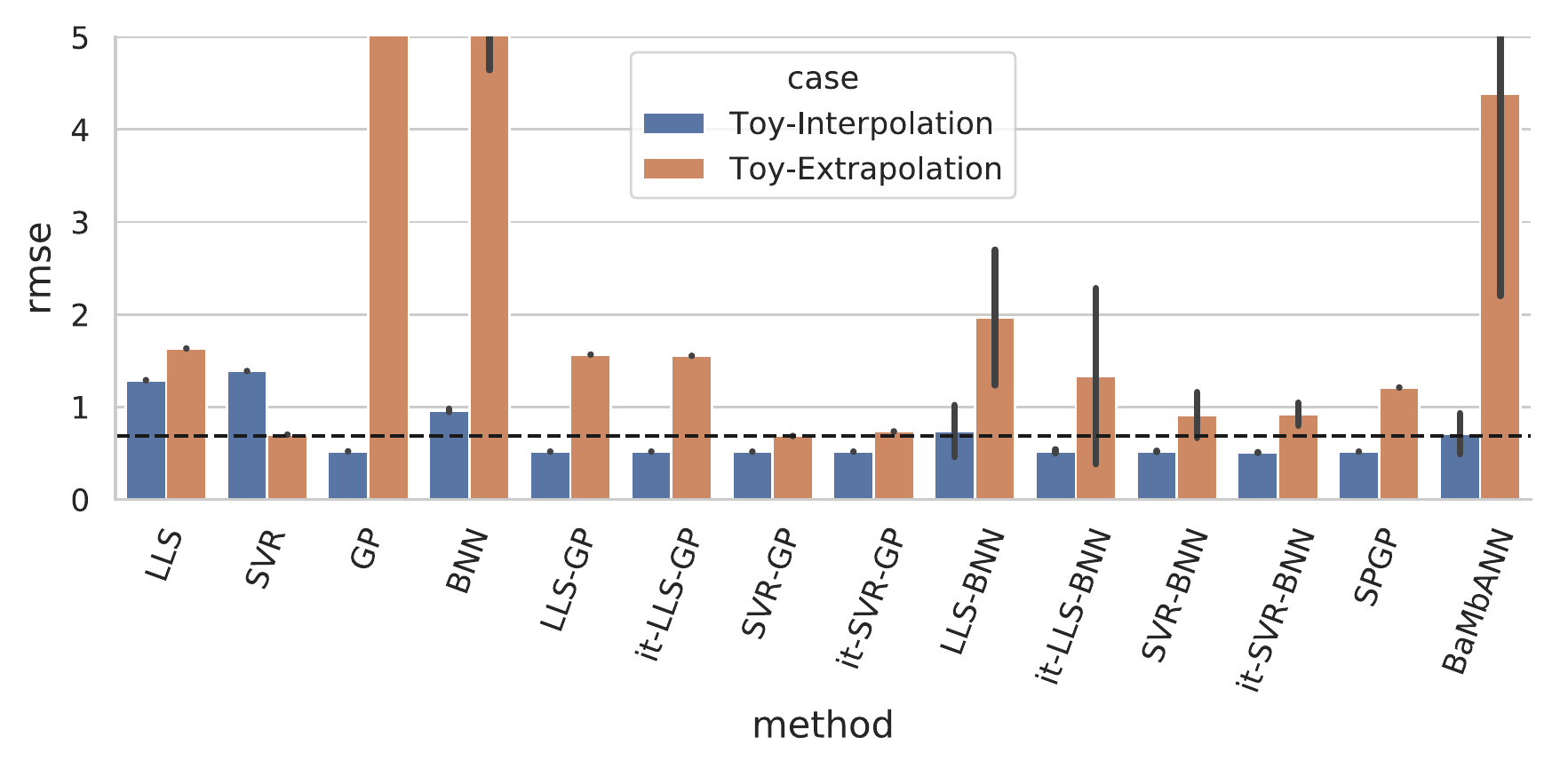}
  \caption{RMSE results for \toy{} test cases. Marked with the horizontal black line is the best extrapolation performance, obtained by \svrgp{} and \svr{}. Error bars illustrate the standard deviation obtained over five independent training runs of the algorithm.}
  \label{fig:toy_rmse}
\end{figure}

The \toy{} scenario marks an idealized version of data which in general follows a global parametric model, but has a local (in input-space) model mismatch. Accurate interpolation requires a non-parametric fit where data is available. Accurate extrapolation requires to extract the right coefficients for the parametric model in presence of the local model mismatch. In Figure \ref{fig:toy_rmse}, the obtained RMSE errors for prediction on the interpolation and extrapolation test set are shown.

As expected, most methods involving a non-parametric part (except some variants involving \bnn{}s) obtain optimal interpolation errors, this holds also for the solely non-parametric \gp{}. While solely parametric methods (\lls{}, \svr{}) obtain higher error in the interpolation setting, they outperform solely non-parametric methods in the extrapolation case by a large margin. If robust model identification is used (\svr{}), extrapolation performance is near optimal. 
Sequential semi-parametric methods based on the \svr{} ((it-)\svrgp{}, (it-)\svrbnn{}) obtain minimal inter- and extrapolation errors within a single model, followed by the joint semi-parametric model \spgp{} and sequential semi-parametric methods based on more outlier-sensitive \lls{} estimation ((it-)\llsgp{}, (it-)\llsbnn{}). \bambann{} fails to identify the model coefficients precise enough to obtain low extrapolation errors and the fit varies strongly with independent repetitions.

\subsection{\vda{} Scenario}

\begin{figure}[htbp]
  \centering
  \includegraphics[width=.5\textwidth]{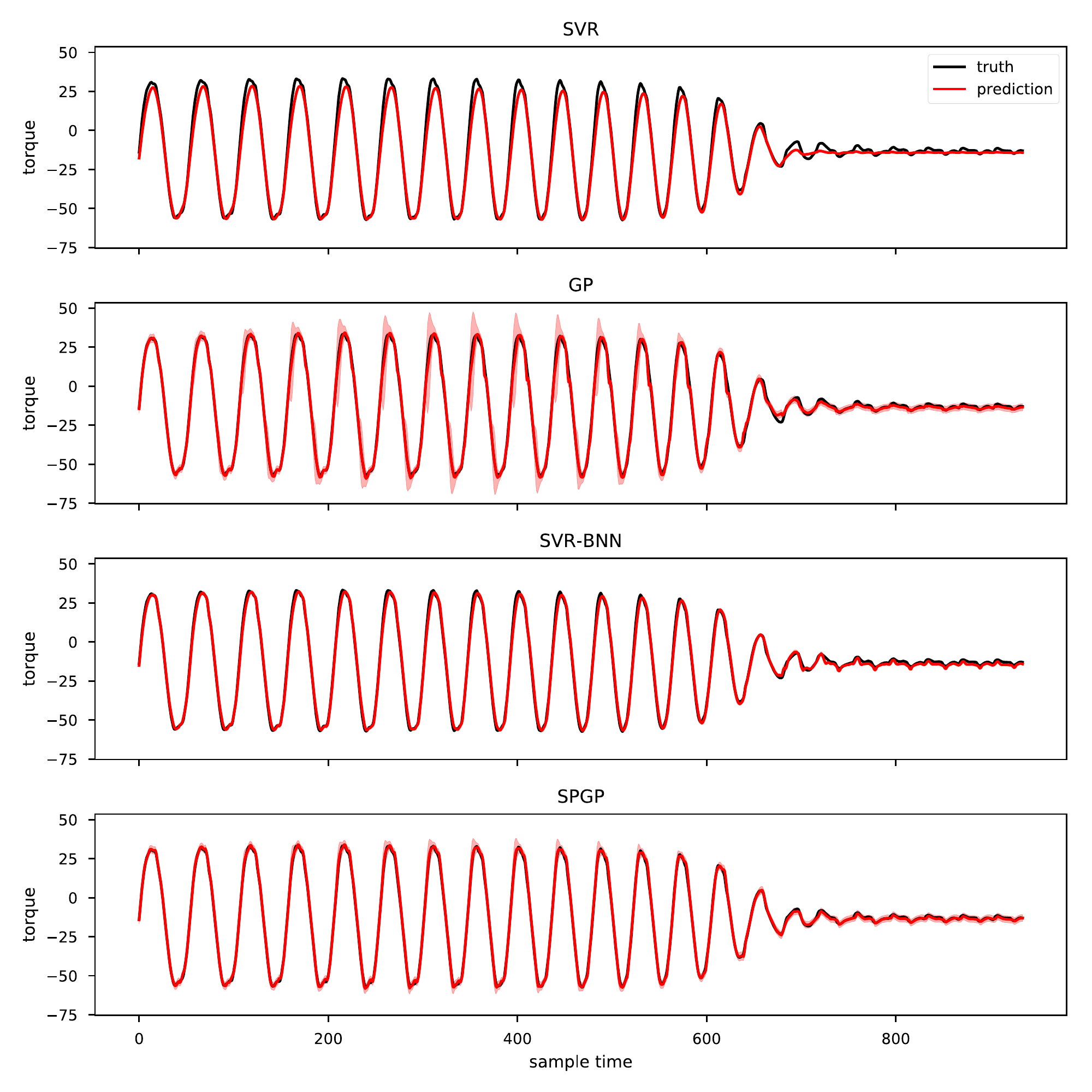}
  \caption{Prediction results for selected methods on the \arvda{} scenario. Shaded area depicts $2\sigma$ prediction uncertainty. \spgp{} has lowest RMSE error.}
  \label{fig:viaarpreds}
\end{figure}

As the \vda{} data set is based on real data, it is difficult to say what exactly causes a mismatch between the parametric model and the data.
As a way of quantifying it, solely parametric methods identify the model with a remaining RMSE error of $\approx4~\mathrm{nm}$ (\lls{}, \svr{}).
As mentioned before, we prepared an instantaneous as well as an auto-regressive setting (\invda{}, \arvda{}) and all methods were evaluated on both.
Example predictions along the test trajectory for the auto-regressive case are illustrated in Figure \ref{fig:viaarpreds}.
The fit for the \svr{} shows that the model by its own is insufficient to capture the peak torques during the chirp motion and does not model the torques during fade-out at the end of the trajectory at all.
Performance in terms of RMSE and negative log-likelihood (NLLH) for both cases are summarized in Figures \ref{fig:via_rmse} and \ref{fig:via_nllh}.

Following Figure \ref{fig:via_rmse}, in the instantaneous setting, the solely non-parametric methods perform similar or worse than the solely parametric methods while the joint semi-parametric methods perform slightly better (\spgp{}, \bambann{}).
The best fit for the \invda{} case in terms of RMSE error is obtained by sequential semi-parametric combinations with a \bnn{} ((it-)\llsbnn{}, (it-)\svrbnn{}), combinations with a \gp{} underperform consistently.
Interestingly, when comparing the negative log-likelihood of the test data for the different methods, all methods with \gp{}s outperform combinations with \bnn{}s by a margin (Figure \ref{fig:via_nllh}).
Many of our experiments showed this behaviour.
In general, we observed that the trained \bnn{}s (and variants thereof) are often overconfident in their prediction which manifests in a worse NLLH score.

As we knew from the test-bed that some physical effects are present which cannot be captured using instantaneous measurements (such as stick-slip friction), it was interesting to see by how much prediction performance improves for the auto-regressive case.
For \arvda{}, we still use the same parametric model which is defined only on the current time-step's data hence \lls{} and \svr{} perform exactly the same.
In the new input space, the non-parametric \gp{} can make much more sense of the data.
Adding a parametric estimator (e.g. (it-)\svrgp{}) does not improve performance further though.
This is interesting as the smallest RMSE error for the \arvda{} case is actually achieved by a semi-parametric approach (\spgp{}) which outperforms all other (sequential or not) semi-parametric methods on this data set.
It also achieves lowest NLLH performance.
\spgp{} in this case estimated the two model coefficients to $K\approx866$ and $D\approx28$ and its optimization was started from a rough guess of $\hat{K} = 300$ and $\hat{D}=20$.
The importance of the start coefficients can be seen when comparing \spgp{} to \spgp{}\_from\_zeros for which the optimization was started from $\hat{K} = \hat{D} = 0.0$ and ended with the physically impossible estimate of $K\approx1460, D\approx-72$ accompanied by worse RMSE performance than the best methods on \invda{} data only.

\begin{figure}[htbp]
  \centering
  \includegraphics[width=.5\textwidth]{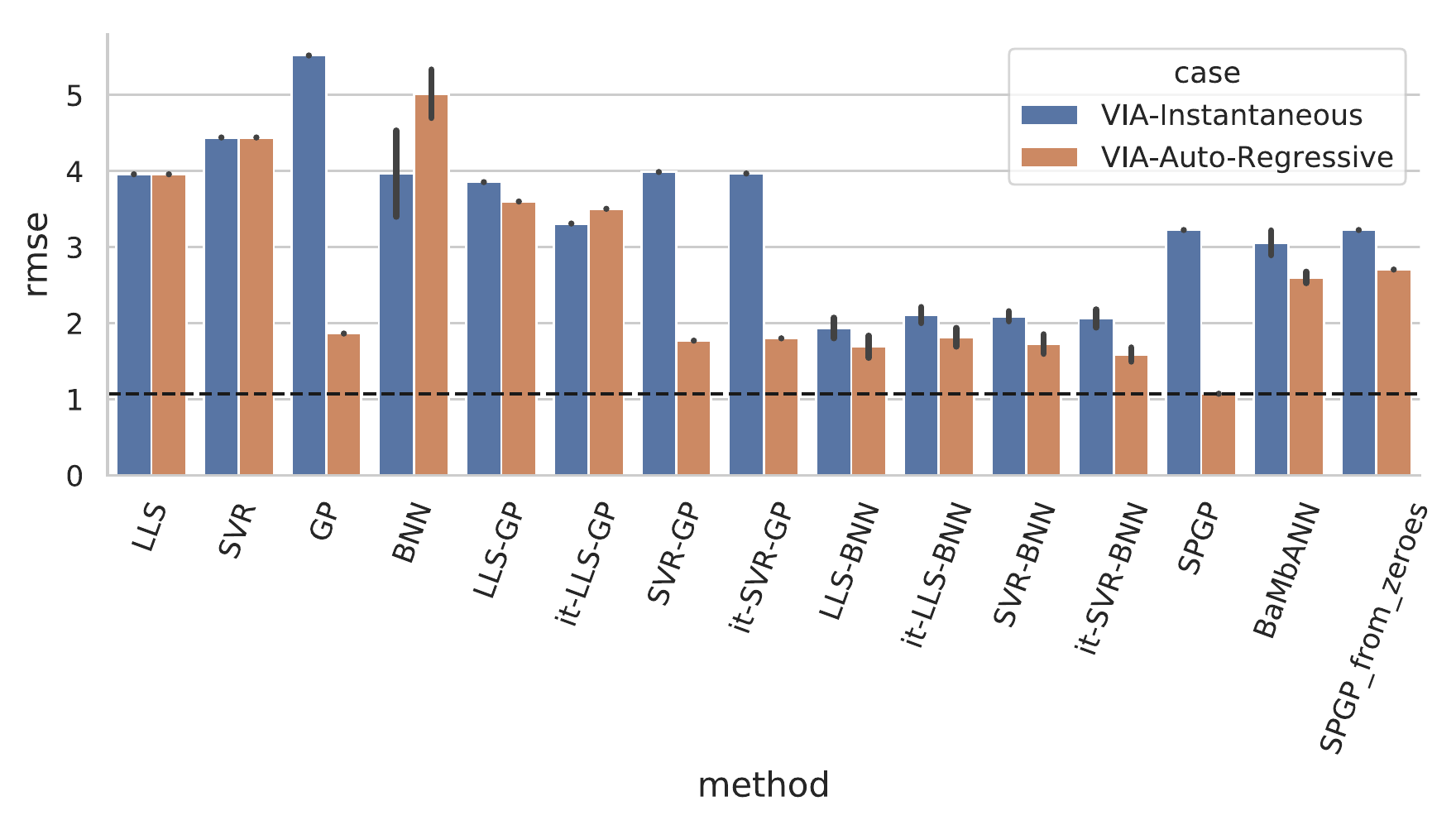}
  \caption{RMSE results for \vda{} test cases. Marked with the horizontal black line is the best auto-regressive performance, obtained by \spgp{}. Error bars illustrate the standard deviation obtained over five independent training runs of the algorithm.}
  \label{fig:via_rmse}
\end{figure}

\begin{figure}[htbp]
  \centering
  \includegraphics[width=.5\textwidth]{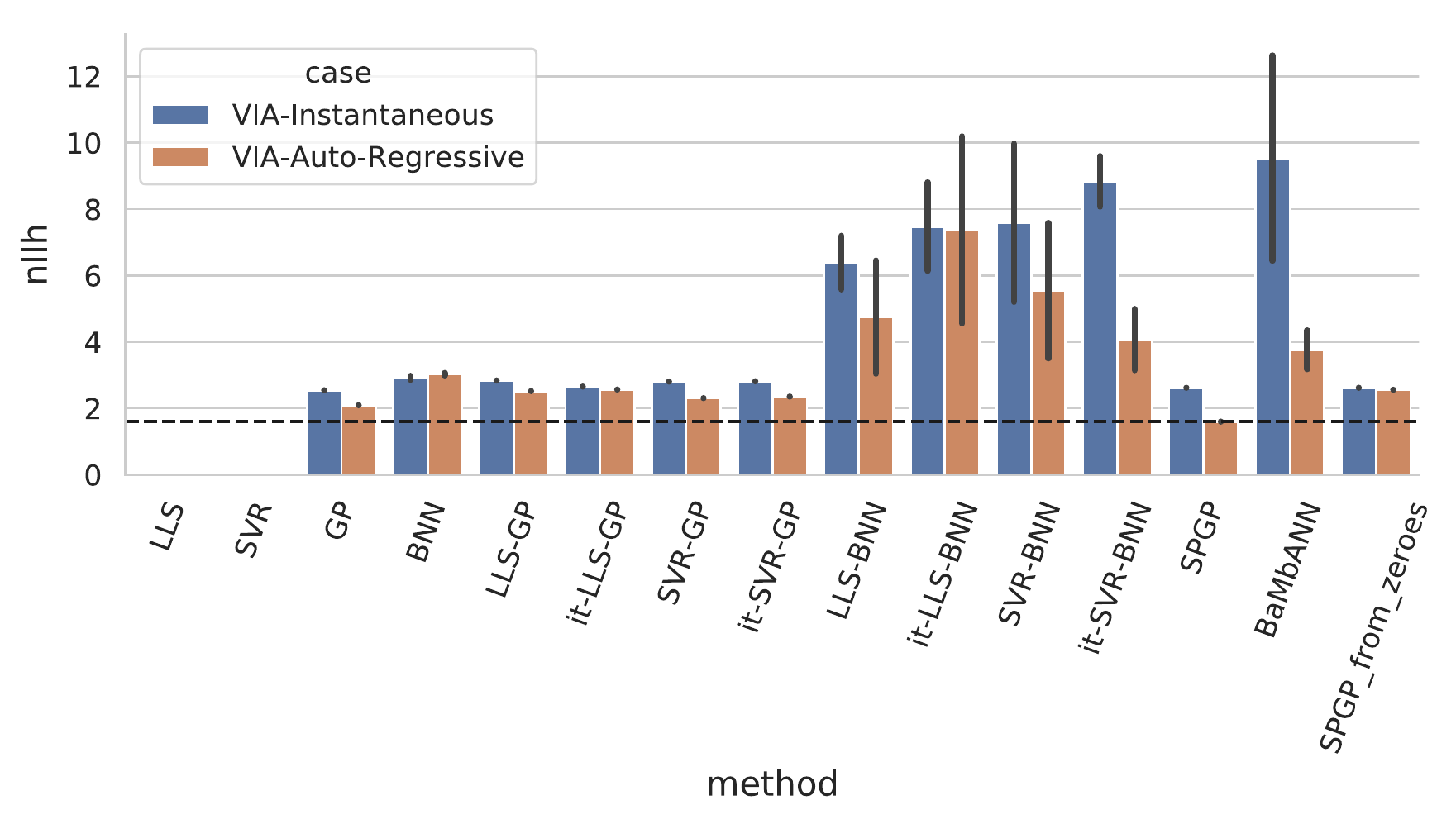}
  \caption{NLLH (negative log-likelihood) results for \vda{} test cases. Marked with the horizontal black line is the best auto-regressive performance, obtained by \spgp{}. Error bars illustrate the standard deviation obtained over five independent training runs of the algorithm.}
  \label{fig:via_nllh}
\end{figure}

\subsection{\simdyn{} Scenarios}

\begin{figure}[htbp]
  \centering
  \includegraphics[width=.5\textwidth]{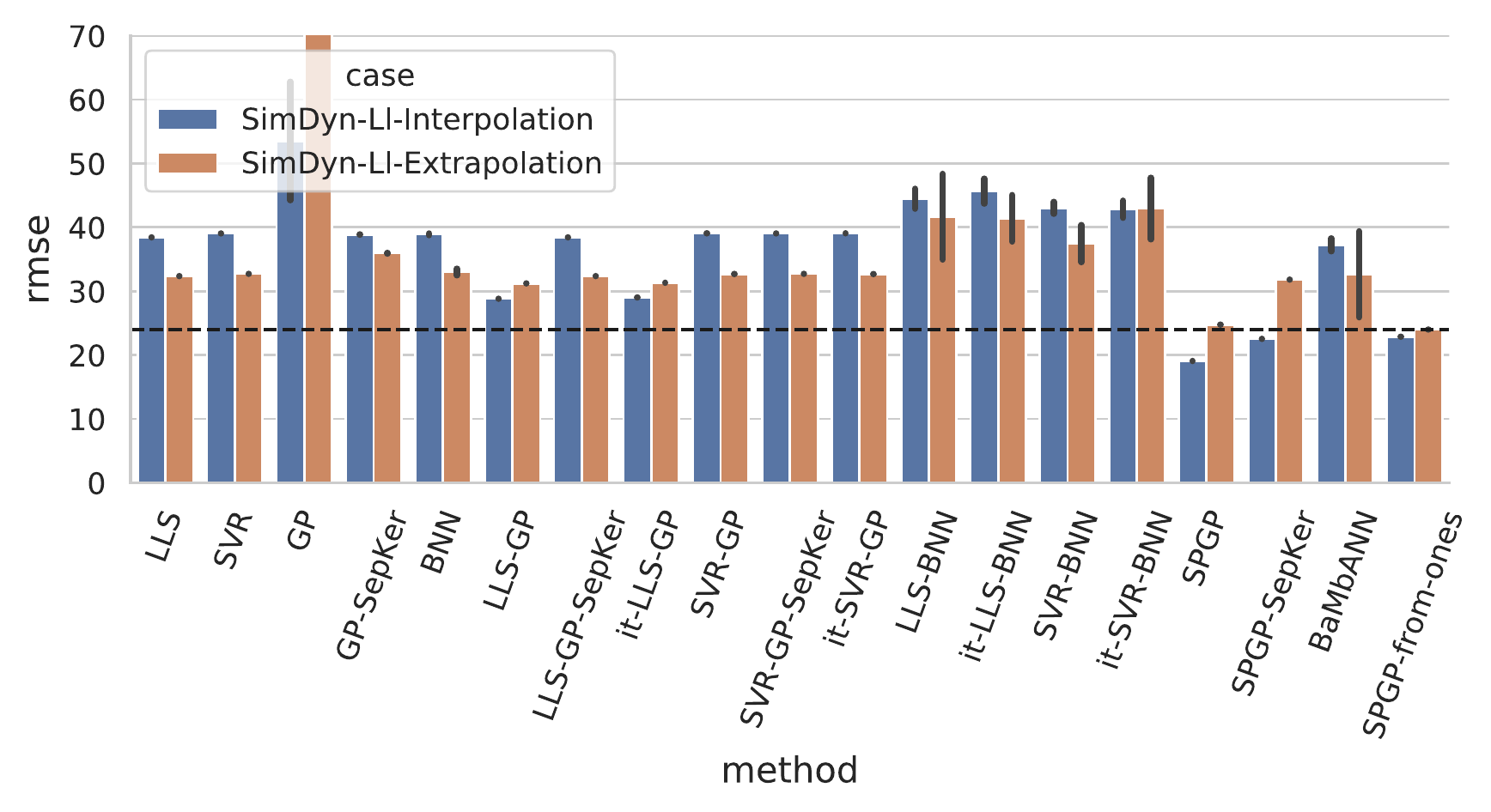}
  \caption{RMSE results for \simdynlocal{} test cases on joint 0. Marked with the horizontal black line is the best extrapolation performance, obtained by \spgp{}\_from\_ones (with a .76-difference to \spgp{}). Error bars illustrate the standard deviation obtained over five independent training runs of the algorithm.}
  \label{fig:simdyn_ll_rmse}
\end{figure}

\begin{figure}[htbp]
  \centering
  \includegraphics[width=.5\textwidth]{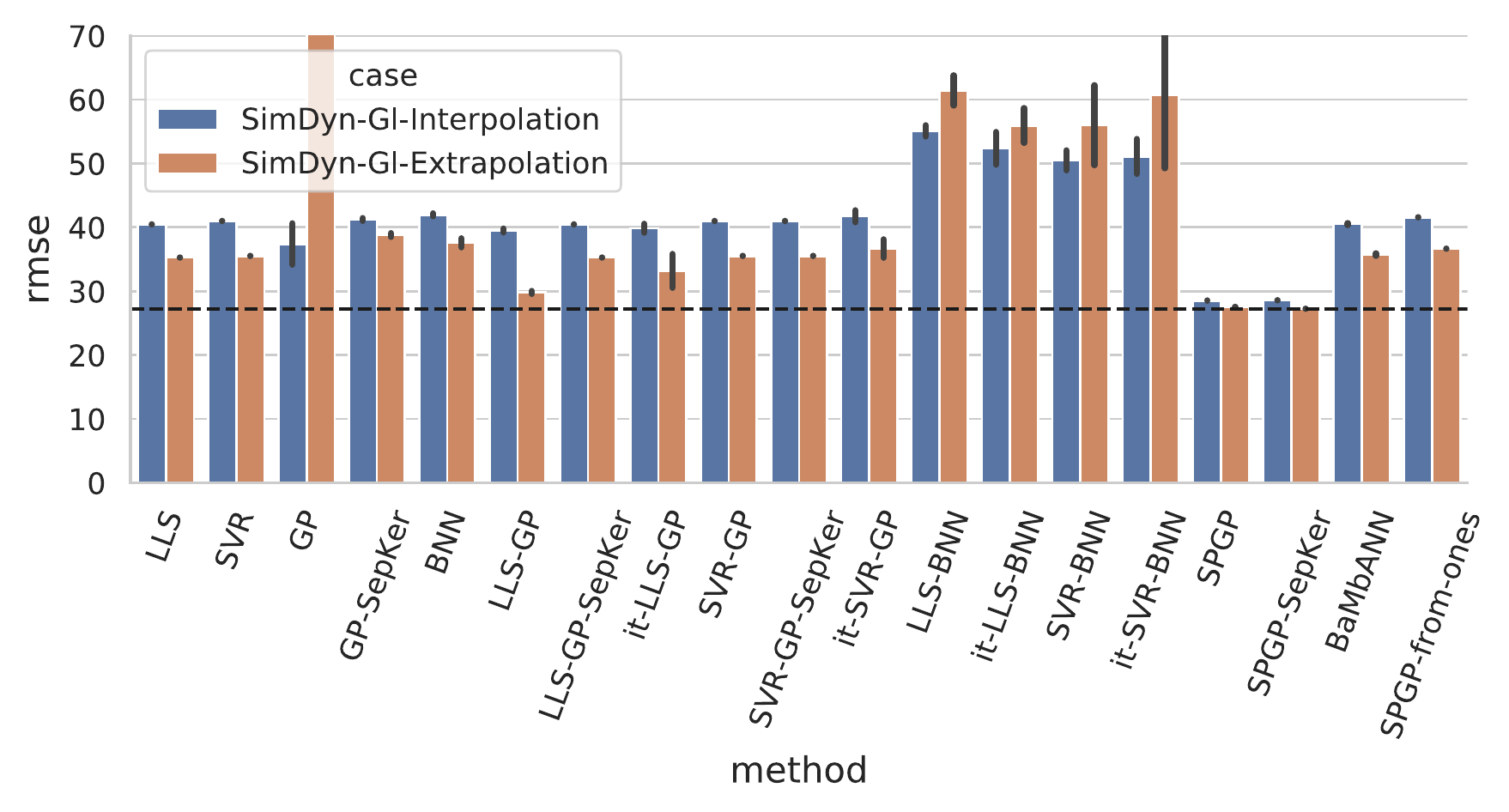}
  \caption{RMSE results for \simdynglobal{} test cases on joint 0. Marked with the horizontal black line is the best extrapolation performance, obtained by \spgp{}\_SepKer (with a .28-difference to \spgp{}). Error bars illustrate the standard deviation obtained over five independent training runs of the algorithm.}
  \label{fig:simdyn_gl_rmse}
\end{figure}

\begin{figure}[htbp]
  \centering
  \includegraphics[width=.5\textwidth]{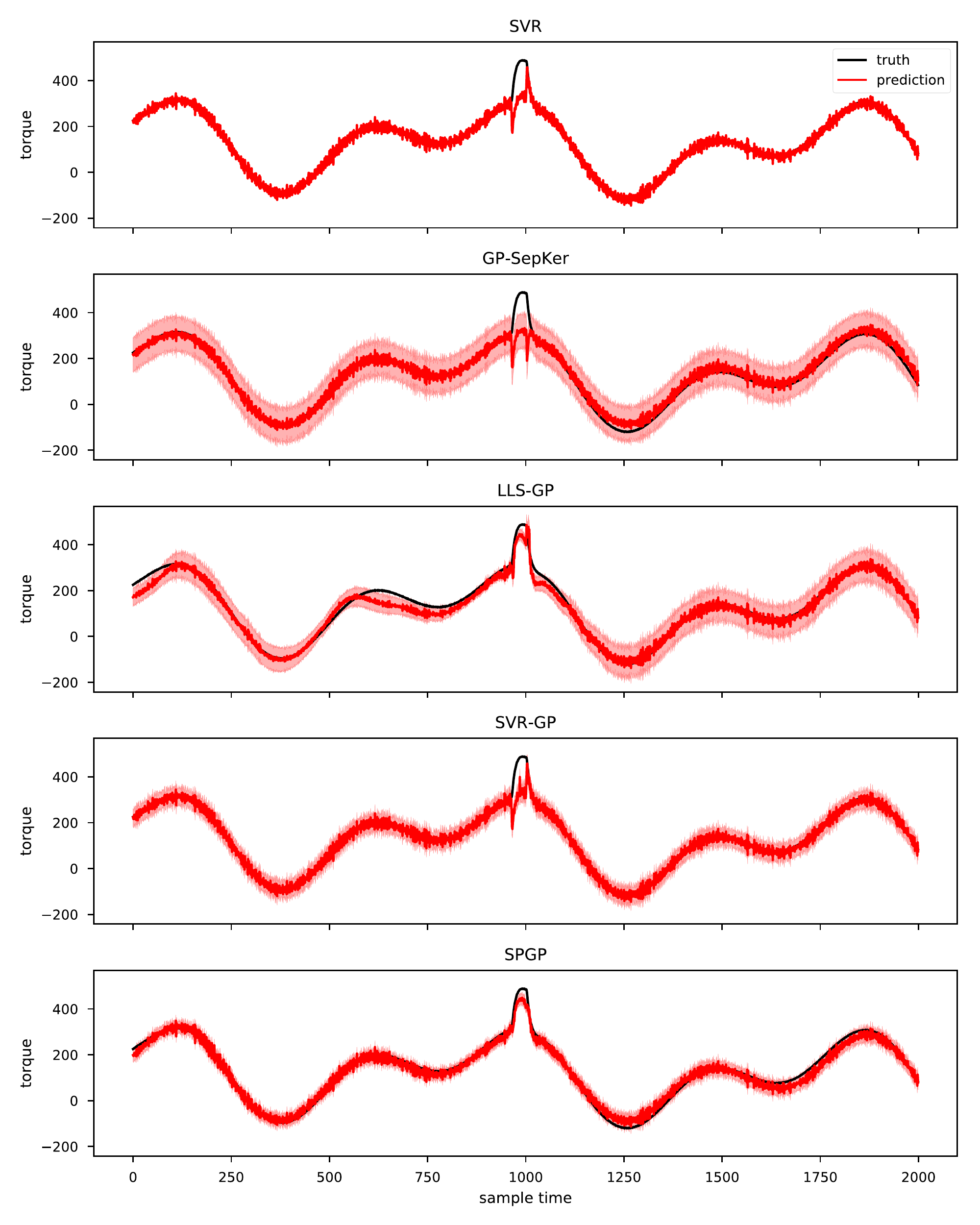}
  \caption{Prediction results for selected methods on the \simdynlocalextra{} scenario and joint 0. Shaded area depicts $2\sigma$ prediction uncertainty. \spgp{} has lowest RMSE error. The peak in required torque prediction at around sample time $\approx 1000$ is due to the local model-mismatch friction on joint 0. The noisy nature of the predictions is due to obtaining $\dot{q},\ddot{q}$ from simulation via a numeric differentiation, leading to ``input noise'' for the learned models.}
  \label{fig:simdynlocalextrapreds}
\end{figure}

For the simulated robot dynamics test scenario, as presented before, we distinguish between the cases i) local model mismatch (first joint only, \simdynlocal{}) and ii) global-with-local model mismatch (\simdynglobal{}).
As this is a multi-target scenario (three joint torques need to be predicted), we trained \gp{}s and \spgp{}s in two flavours.
One in which kernel hyper-parameters (and in case of \spgp{} mean function parameters) are shared among all three output dimensions and one in which they are not (variants with suffix \emph{-SepKer} for ``separate kernels'').
A qualitative impression of the predictions for selected methods on the extrapolation trajectory for the \simdynlocal{} mismatch is presented in Figure \ref{fig:simdynlocalextrapreds}.
Only few methods reliably model the torque peak required to overcome the local friction on joint 0.
RMSE errors for joint 0 and all tested methods are illustrated in Figures \ref{fig:simdyn_ll_rmse} and \ref{fig:simdyn_gl_rmse}.
For joint 0, the errors in both settings (\simdynlocal{} and \simdynglobal{}) are mostly determined by the methods ability to model the strong local friction behavior as the error introduced by the local mismatch is much higher than error from the global ``mismatch ripple torque''.

In general, with both test scenarios we found that only few methods stand out compared to the solely parametric solutions.
For \simdynlocal{}, the only methods which are able to capture the local friction behavior are the \spgp{} variants and to some degree the sequential semi-parametric methods (it-)\llsgp{}.
For the \spgp{} fit, we found that in the case of additional global model mismatch, the initialization of the parametric coefficients plays a role for the optimization (cf. Figure \ref{fig:simdyn_gl_rmse} \spgp{}\_from\_ones vs. \spgp{}, were optimization for \spgp{} is started from the values as described in Table \ref{tab:simdyncoeffs} and optimization for \spgp{}\_from\_ones is started from a vector of ones $\boldsymbol{1} \in \mathbb{R}^{17}$).
\gp{}s trained with a shared kernel between the outputs fail badly at extrapolating and when trained with separate kernels per output dimension only manage to fit the general robot dynamics but not the local friction phenomena (which results in similar performance than the solely parametric methods).

We also see that in many cases the RMSE error on the extrapolation data set is lower than on the interpolation test data, especially for algorithms which are not good at fitting the local disturbance. We investigated this and found that although inter- and extrapolation trajectories pass through the high-friction area equally often, the extrapolation trajectory passes through the area at a generally higher joint speed which means the (sample) time in which errors for wrong predictions accumulate is shorter which in turn leads to lower RMSE values.

\subsection{Discussion and Remarks}

From the experiments we conducted, a few general remarks can be made:
\begin{itemize}
\item \spgp{} when initialized with a reasonable guess for the parametric coefficients provides robust and better estimation performance than all other evaluated methods in terms of RMSE and also NLLH scores. Only in the ``staged'' \toyextra{} scenario, \spgp{} is beaten by a sequential semi-parametric approach and only if robust parametric estimation by means of \svr{} is used.
\item Training of \bnn{}s to obtain good estimation performance is not easy and might require more investigation into what hyper-parameters to use. The combination of parametric with dense layers in the way we did was therefore seldom fruitful (especially compared to \spgp{}). Also, the obtained uncertainty estimates of the \bnn{}-based approaches are often overconfident, limiting their use in practice.
\item Comparing \llsgp{} and \svrgp{} as naive approaches to sequential semi-parametric model estimation, we would recommend to use the more robust \svr{} to obtain a more robust model calibration.
\item Iterating parametric and non-parametric fitting in the style of \itllsgp, \itsvrgp, \ldots does not improve performance.
\end{itemize}

\section{Related Work}
\label{sec:rw}

The literature in the field of model learning for robot control is vast \cite{nguyen-tuong2011model, sigaud2011line}.
In the more recent survey by Chatzilygeroudis et al. \cite{chatzilygeroudis2018survey}, chapter 5.2 is dedicated to learning robot models for control using ``priors on the dynamics'' and it lists several references to other works implementing such strategies.
As our focus is on joint semi-parametric model learning with an analytic model as basis, we think the following literature is most relevant.

Quite some papers implemented and evaluated Gaussian Processes (GP) using a parametric model as mean function in some form \cite{nguyen-tuong2010using, wu2012semi, krishnamoorthi2018model, chatzilygeroudis2018using}.
Closely related to our formulation of \spgp{} are the variants described in \cite{nguyen-tuong2010using} and \cite{wu2012semi}.
Both assume a linear-in-parameters mean function and apply the equations described in \cite[chapter~2.7]{rasmussen2003gaussian} to marginalize out the parametric model coefficients.
In contrast, we simply optimize over model coefficients the same way as over kernel hyper-parameters which would in principle allow for non-linear-in-parameters parametric models in our case but not theirs (but all described models in this letter are actually linear-in-parameters).
In \cite{krishnamoorthi2018model} and as the comparison variant ``RBD Mean'' in \cite{nguyen-tuong2010using}, a parametric model with fixed, predetermined coefficients is used.
Both papers mention linear identification techniques for estimating the coefficients and thus the results for these methods should be equivalent or similar to our baseline method \llsgp{}.
Our baseline \svrgp{} provides an interesting twist on that approach by evaluating a simple way to robustify the parametric model identification against outliers and model-vs-data mismatch.
In \cite{chatzilygeroudis2018using}, no assumptions are made regarding the structure of the mean function.
As they neither require the parametric model to be linear-in-parameters nor differentiable with respect to the parameters, they have to resort to a two-staged, iterative model learning procedure involving a black-box, derivative-free optimization for the model parameters and regular gradient-based optimization for \gp{} kernel parameters.
As the models considered in this letter are differentiable with respect to the model parameters, we can use gradient-based optimization jointly for all parameters and did not evaluate such black-box optimization techniques. Nevertheless, their results look promising for cases where a parametric model gets more complicated.

An approach to online sequential model calibration with non-parametric residual learning is presented in \cite{camoriano2016incremental}.
At the core of their approach lies the well known recursive regularized least squares algorithm.
At every time step it is used to incrementally refine the parametric model coefficients via a linear least squares formulation (assumes linear-in-parameters parametric model).
Subsequently, also at every time step, the same algorithm is used to incrementally fit a non-parametric, non-linear regressor for the residual dynamics. This non-parametric regressor is related to incremental sparse spectrum Gaussian process regression \cite{gijsberts2013real} and therefore the batch result for their algorithm should be equivalent to our sequential semi-parametric variant \llsgp{}.

A more involved approach to parametric model identification than simple \lls{} or \svr{} is presented in \cite{ting2006bayesian}.
They use a Bayesian approach to fit model parameters to data while being robust to input/output noise, ill-conditioned data, non-identifiable parameters and physically non-plausible parameters.
Such an approach could be used as drop-in replacement for \lls{} or \svr{} in any of our sequential semi-parametric methods.

We did not find any literature on combining parametric models with neural networks as in the proposed \bambann{} method. A neural network approach which can solve system identification problems is presented in \cite{sanchez-gonzalez2018graph}, but they are not using an analytic model of the system as basis.
They provide a mechanism for a (graph) network to infer and (re)use static parameters to describe a system.
They term this implicit system identification, because these parameters may well be related to static, physical properties like mass or inertia, but what they represent is learned/optimized from data hence not interpretable in a straight-forward way.

\section{Conclusion}
\label{sec:conclusion}
In this letter, we compared several parametric, non-parametric and semi-parametric regression methods on three test datasets of increasing complexity and where possible in an inter- as well as extrapolation setting.
We evaluated prediction accuracy in terms of RMSE error as well as the quantification of uncertainty by means of the negative log-likelihood (for methods which provide uncertainty estimates).
Of particular interest to us were semi-parametric modeling approaches and therein the comparison of joint versus sequential semi-parametric methods.
In addition to state-of-the-art (joint) semi-parametric Gaussian process regression, we also developed and evaluated a joint semi-parametric neural network architecture and implemented several sequential semi-parametric approaches combining common parametric with non-parametric models as baselines.
The general findings in terms of accuracy in inter- vs. extrapolation settings for solely parametric and solely non-parametric models were as expected (cf. Table \ref{tab:strenghts}).
For the semi-parametric methods, we see a clear benefit in joint semi-parametric modeling with respect to the sequential semi-parametric baselines.
In all but one case, semi-parametric Gaussian process regression, through its principled approach in integrating a given parametric model, provides the best model accuracy (RMSE) with robust estimates of predicted uncertainty (NLLH).
For our joint semi-parametric neural network approach, we found that while its accuracy \emph{can} be in the regime of sequential semi-parametric approaches, it never matches semi-parametric Gaussian process regression and - over independent training runs - shows much higher variability in the achieved model accuracy than other approaches.
A further investigation into the various involved hyper-parameters for this approach could be valuable.

In general, we think the benefit of joint semi-parametric modeling, especially in extrapolation settings, warrants more research into such modeling approaches.



\bibliographystyle{./style/IEEEtran}
\bibliography{bibliography}

\end{document}


\newcommand{\sysid}{system identification}

\newcommand{\toy}{\textsc{Toy}}
\newcommand{\toyinter}{\textsc{Toy-Interpolation}}
\newcommand{\toyextra}{\textsc{Toy-Extrapolation}}
\newcommand{\vda}{\textsc{VIA}}
\newcommand{\invda}{\textsc{VIA-Instantaneous}}
\newcommand{\arvda}{\textsc{VIA-Auto-Regressive}}
\newcommand{\simdyn}{\textsc{SimDyn}}
\newcommand{\simdynglobal}{\textsc{SimDyn-Gl}}
\newcommand{\simdynglobalinter}{\textsc{SimDyn-Gl-Interpolation}}
\newcommand{\simdynglobalextra}{\textsc{SimDyn-Gl-Extrapolation}}
\newcommand{\simdynlocal}{\textsc{SimDyn-Ll}}
\newcommand{\simdynlocalinter}{\textsc{SimDyn-Ll-Interpolation}}
\newcommand{\simdynlocalextra}{\textsc{SimDyn-Ll-Extrapolation}}

\newcommand{\mytodo}[1]{\textbf{\textcolor{red}{TODO: #1}}}

\newcommand{\spgp}{\textsc{SPGP}}
\newcommand{\gp}{\textsc{GP}}
\newcommand{\bambann}{\textsc{BaMbANN}}
\newcommand{\bnn}{\textsc{BNN}}
\newcommand{\lls}{\textsc{LLS}}
\newcommand{\llsgp}{\textsc{LLS-GP}}
\newcommand{\llsbnn}{\textsc{LLS-BNN}}
\newcommand{\itllsgp}{it-\textsc{LLS-GP}}
\newcommand{\itllsbnn}{it-\textsc{LLS-BNN}}
\newcommand{\svr}{\textsc{SVR}}
\newcommand{\svrgp}{\textsc{SVR-GP}}
\newcommand{\svrbnn}{\textsc{SVR-BNN}}
\newcommand{\itsvrgp}{it-\textsc{SVR-GP}}
\newcommand{\itsvrbnn}{it-\textsc{SVR-BNN}}

\begin{acronym}[SPML]
\acro{RL}{reinforcement learning}
\acro{GP}{Gaussian process}
\acro{NN}{neural network}
\end{acronym}

\maketitle

\section{Predictions}

\begin{figure*}[htbp]
  \centering
  \includegraphics[width=\textwidth]{images-bin/toy_exp_interp_w_noise-png.png}
  \caption{Prediction results for the \toyinter{} test data. Columns show different independent training repetitions for each method. Rows show all evaluated methods. In red, the prediction with a shaded $2\sigma$ (were available) is shown. Black shows ground truth data. Blue the training data.}
  \label{fig:toyinter}
\end{figure*}

\begin{figure*}[htbp]
  \centering
  \includegraphics[width=\textwidth]{images-bin/toy_exp_extrap_w_noise-png.png}
  \caption{Prediction results for the \toyextra{} test data. Columns show different independent training repetitions for each method. Rows show all evaluated methods. In red, the prediction with a shaded $2\sigma$ (were available) is shown. Black shows ground truth data. Blue the training data.}
  \label{fig:toyextra}
\end{figure*}

\begin{figure*}[htbp]
  \centering
  \includegraphics[width=\textwidth]{images-bin/vda_instant_preds-png.png}
  \caption{Prediction results for the \invda{} test data. Columns show different independent training repetitions for each method. Rows show all evaluated methods. In red, the prediction with a shaded $2\sigma$ (were available) is shown. Black shows ground truth data.}
  \label{fig:invda}
\end{figure*}

\begin{figure*}[htbp]
  \centering
  \includegraphics[width=\textwidth]{images-bin/vda_ar_preds-png.png}
  \caption{Prediction results for the \arvda{} test data. Columns show different independent training repetitions for each method. Rows show all evaluated methods. In red, the prediction with a shaded $2\sigma$ (were available) is shown. Black shows ground truth data.}
  \label{fig:arvda}
\end{figure*}

\begin{figure*}[htbp]
  \centering
  \includegraphics[width=\textwidth]{images-bin/simdyn_preds_exp_interp_SENSING_dist_local-png.png}
  \caption{Prediction results for the \simdynlocalinter{} test data. Illustrated are the torque predictions for joint 0 on the last two seconds (2000 samples) of the data. Columns show different independent training repetitions for each method. Rows show all evaluated methods. In red, the prediction with a shaded $2\sigma$ (were available) is shown. Black shows ground truth data.}
  \label{fig:simdynlocalinter}
\end{figure*}

\begin{figure*}[htbp]
  \centering
  \includegraphics[width=\textwidth]{images-bin/simdyn_preds_exp_extrap_SENSING_dist_local-png.png}
  \caption{Predictions results for the \simdynlocalextra{} test data. Illustrated are the torque predictions for joint 0 on the last two seconds (2000 samples) of the data. Columns show different independent training repetitions for each method. Rows show all evaluated methods. In red, the prediction with a shaded $2\sigma$ (were available) is shown. Black shows ground truth data.}
  \label{fig:simdynlocalextra}
\end{figure*}

\begin{figure*}[htbp]
  \centering
  \includegraphics[width=\textwidth]{images-bin/simdyn_preds_exp_interp_SENSING-png.png}
  \caption{Predictions results for the \simdynglobalinter{} test data. Illustrated are the torque predictions for joint 0 on the last two seconds (2000 samples) of the data. Columns show different independent training repetitions for each method. Rows show all evaluated methods. In red, the prediction with a shaded $2\sigma$ (were available) is shown. Black shows ground truth data.}
  \label{fig:simdynglobalinter}
\end{figure*}

\begin{figure*}[htbp]
  \centering
  \includegraphics[width=\textwidth]{images-bin/simdyn_preds_exp_extrap_SENSING-png.png}
  \caption{Predictions results for the \simdynglobalextra{} test data. Illustrated are the torque predictions for joint 0 on the last two seconds (2000 samples) of the data. Columns show different independent training repetitions for each method. Rows show all evaluated methods. In red, the prediction with a shaded $2\sigma$ (were available) is shown. Black shows ground truth data.}
  \label{fig:simdynglobalextra}
\end{figure*}

\FloatBarrier
\section{Complete Aggregated Quantitative Results}

\begin{table*}[htbp]
\centering
\caption {\label{tab:toyinter} Results \toy{} - Interpolation. Statistics are built over five independent executions of each algorithm.} 
\begin{tabular}{l||ll|ll}
Method & \multicolumn{4}{c}{\toy{} - Interpolation} \\
\hline
       & \multicolumn{2}{c|}{RMSE} & \multicolumn{2}{c}{NLLH} \\ 
\cline{2-5}
       &  mean $\pm$ std. & min / max & mean $\pm$ std. & min / max \\
\hline
\begin{filecontents*}{./csvs/toy_interp_w_noise.csv}
algo,rmsemean,rmsestd,rmseamin,rmseamax,nllhmean,nllhstd,nllhamin,nllhamax
LLS,1.29,0.00,1.29,1.29,inf,-,inf,inf
LLS-GP,0.52,0.00,0.52,0.52,0.76,0.00,0.76,0.76
it-LLS-GP,0.52,0.00,0.52,0.52,0.76,0.00,0.76,0.76
LLS-BNN,0.74,0.31,0.52,1.29,inf,-,2.63,inf
it-LLS-BNN,0.52,0.02,0.51,0.56,9.13,3.44,5.13,12.48
SVR,1.39,0.00,1.39,1.39,inf,-,inf,inf
SVR-GP,0.52,0.00,0.52,0.52,0.76,0.00,0.76,0.76
it-SVR-GP,0.52,0.00,0.52,0.52,0.76,0.00,0.76,0.76
SVR-BNN,0.52,0.01,0.51,0.52,19.23,5.31,14.13,28.25
it-SVR-BNN,0.51,0.00,0.50,0.51,13.51,5.62,4.16,17.35
GP,0.52,0.00,0.52,0.52,0.77,0.00,0.77,0.77
BNN,0.96,0.02,0.95,0.99,2.74,0.32,2.30,3.11
SPGP,0.52,0.00,0.52,0.52,0.76,0.00,0.76,0.76
BaMbANN,0.71,0.25,0.51,0.98,2.62,1.94,0.96,4.98
\end{filecontents*}
\csvreader[head to column names]{./csvs/toy_interp_w_noise.csv}{}
{\\\algo & \rmsemean ~$\pm$ \rmsestd & \rmseamin ~/ \rmseamax & \nllhmean ~$\pm$ \nllhstd & \nllhamin 	~/ \nllhamax}
\end{tabular}
\end{table*}

\begin{table*}[htbp]
\centering
\caption {\label{tab:toyextra} Results \toy{} - Extrapolation. Statistics are built over five independent executions of each algorithm.} 
\begin{tabular}{l||ll|ll}
Method & \multicolumn{4}{c}{\toy{} - Extrapolation} \\
\hline
       & \multicolumn{2}{c|}{RMSE} & \multicolumn{2}{c}{NLLH} \\ 
\cline{2-5}
       &  mean $\pm$ std. & min / max & mean $\pm$ std. & min / max \\
\hline
\begin{filecontents*}{./csvs/toy_extrap_w_noise.csv}
algo,rmsemean,rmsestd,rmseamin,rmseamax,nllhmean,nllhstd,nllhamin,nllhamax
LLS,1.63,0.00,1.63,1.63,inf,-,inf,inf
LLS-GP,1.57,0.00,1.57,1.57,1.87,0.00,1.87,1.87
it-LLS-GP,1.55,0.00,1.55,1.55,1.86,0.00,1.86,1.86
LLS-BNN,1.97,0.82,1.07,2.89,inf,-,16.51,inf
it-LLS-BNN,1.33,1.06,0.72,3.21,24.13,33.51,1.81,83.15
SVR,0.70,0.00,0.70,0.70,inf,-,inf,inf
SVR-GP,0.69,0.00,0.69,0.69,1.34,0.00,1.34,1.34
it-SVR-GP,0.73,0.00,0.73,0.73,1.36,0.00,1.36,1.36
SVR-BNN,0.91,0.28,0.61,1.35,13.86,3.16,9.73,16.94
it-SVR-BNN,0.92,0.14,0.79,1.15,16.01,10.79,4.82,27.98
GP,17.66,0.00,17.66,17.66,8.19,0.00,8.19,8.19
BNN,5.61,1.08,4.66,6.78,22.49,5.88,16.81,29.19
SPGP,1.21,0.00,1.21,1.21,1.59,0.00,1.59,1.59
BaMbANN,4.39,2.45,2.10,7.25,26.47,31.34,4.51,76.45
\end{filecontents*}
\csvreader[head to column names]{./csvs/toy_extrap_w_noise.csv}{}
{\\\algo & \rmsemean ~$\pm$ \rmsestd & \rmseamin ~/ \rmseamax & \nllhmean ~$\pm$ \nllhstd & \nllhamin 	~/ \nllhamax}
\end{tabular}
\end{table*}

\begin{table*}[htbp]
\centering
\caption {\label{tab:invda} Results \invda{}. Statistics are built over five independent executions of each algorithm.} 
\begin{tabular}{l||ll|ll}
Method & \multicolumn{4}{c}{\invda{}} \\
\hline
       & \multicolumn{2}{c|}{RMSE} & \multicolumn{2}{c}{NLLH} \\ 
\cline{2-5}
       &  mean $\pm$ std. & min / max & mean $\pm$ std. & min / max \\
\hline
\begin{filecontents*}{./csvs/vda_instant.csv}
algo,rmsemean,rmsestd,rmseamin,rmseamax,nllhmean,nllhstd,nllhamin,nllhamax,outputdim
LLS,3.96,0.00,3.96,3.96,inf,-,inf,inf,0
LLS-GP,3.85,0.00,3.85,3.85,2.83,0.00,2.83,2.83,0
it-LLS-GP,3.31,0.00,3.31,3.31,2.65,0.00,2.65,2.65,0
LLS-BNN,1.94,0.15,1.68,2.08,6.39,0.90,5.11,7.64,0
it-LLS-BNN,2.11,0.12,1.95,2.24,7.47,1.49,5.99,9.48,0
SVR,4.44,0.00,4.44,4.44,inf,-,inf,inf,0
SVR-GP,3.99,0.00,3.99,3.99,2.80,0.00,2.80,2.80,0
it-SVR-GP,3.97,0.00,3.97,3.97,2.81,0.00,2.81,2.81,0
SVR-BNN,2.09,0.07,2.03,2.22,7.58,2.66,5.22,12.00,0
it-SVR-BNN,2.06,0.13,1.89,2.24,8.83,0.85,7.67,9.97,0
GP,5.52,0.00,5.52,5.52,2.54,0.00,2.54,2.54,0
BNN,3.96,0.63,3.26,4.76,2.91,0.07,2.84,3.01,0
SPGP,3.22,0.00,3.22,3.22,2.62,0.00,2.62,2.62,0
BaMbANN,3.06,0.18,2.92,3.37,9.54,3.46,6.05,15.25,0
SPGP-from-zeroes,3.22,0.00,3.22,3.22,2.62,0.00,2.62,2.62,0
\end{filecontents*}
\csvreader[head to column names]{./csvs/vda_instant.csv}{}
{\\\algo & \rmsemean ~$\pm$ \rmsestd & \rmseamin ~/ \rmseamax & \nllhmean ~$\pm$ \nllhstd & \nllhamin 	~/ \nllhamax}
\end{tabular}
\end{table*}

\begin{table*}[htbp]
\centering
\caption {\label{tab:arvda} Results \arvda{}. Statistics are built over five independent executions of each algorithm.} 
\begin{tabular}{l||ll|ll}
Method & \multicolumn{4}{c}{\arvda{}} \\
\hline
       & \multicolumn{2}{c|}{RMSE} & \multicolumn{2}{c}{NLLH} \\ 
\cline{2-5}
       &  mean $\pm$ std. & min / max & mean $\pm$ std. & min / max \\
\hline
\begin{filecontents*}{./csvs/vda_ar.csv}
algo,rmsemean,rmsestd,rmseamin,rmseamax,nllhmean,nllhstd,nllhamin,nllhamax,outputdim
LLS,3.96,0.00,3.96,3.96,inf,-,inf,inf,0
LLS-GP,3.60,0.00,3.60,3.60,2.52,0.00,2.52,2.52,0
it-LLS-GP,3.50,0.00,3.50,3.50,2.56,0.00,2.56,2.56,0
LLS-BNN,1.69,0.16,1.47,1.90,4.74,1.91,2.55,7.18,0
it-LLS-BNN,1.81,0.14,1.73,2.05,7.37,3.15,5.54,12.95,0
SVR,4.44,0.00,4.44,4.44,inf,-,inf,inf,0
SVR-GP,1.77,0.00,1.77,1.77,2.30,0.00,2.30,2.30,0
it-SVR-GP,1.80,0.00,1.80,1.80,2.35,0.00,2.35,2.35,0
SVR-BNN,1.72,0.14,1.57,1.95,5.54,2.28,3.00,9.19,0
it-SVR-BNN,1.59,0.10,1.45,1.72,4.07,1.03,3.08,5.63,0
GP,1.86,0.00,1.86,1.86,2.09,0.00,2.09,2.09,0
BNN,5.02,0.36,4.42,5.31,3.02,0.04,2.98,3.09,0
SPGP,1.07,0.00,1.07,1.07,1.59,0.00,1.59,1.59,0
BaMbANN,2.60,0.08,2.52,2.74,3.76,0.65,3.39,4.92,0
SPGP-from-zeroes,2.70,0.00,2.70,2.70,2.56,0.00,2.56,2.56,0
\end{filecontents*}
\csvreader[head to column names]{./csvs/vda_ar.csv}{}
{\\\algo & \rmsemean ~$\pm$ \rmsestd & \rmseamin ~/ \rmseamax & \nllhmean ~$\pm$ \nllhstd & \nllhamin 	~/ \nllhamax}
\end{tabular}
\end{table*}

\begin{table*}
\centering
\caption {\label{tab:simdynglobalinter} Results \simdynglobalinter{}, joint 0. Statistics are built over five independent executions of each algorithm.} 
\begin{tabular}{l||ll|ll}
Method & \multicolumn{4}{c}{\simdynglobalinter{}, joint 0} \\
\hline
     & \multicolumn{2}{c|}{RMSE} & \multicolumn{2}{c}{NLLH} \\ 
\cline{2-5}
     &  mean $\pm$ std. & min / max & mean $\pm$ std. & min / max \\
\hline
\begin{filecontents*}{./csvs/simdyn_stats_exp_interp_SENSING_dim_0.csv}
algo,rmsemean,rmsestd,rmseamin,rmseamax,nllhmean,nllhstd,nllhamin,nllhamax
LLS-simulated,41.43,-,41.43,41.43,inf,-,inf,inf
LLS-prior,42.20,-,42.20,42.20,inf,-,inf,inf
LLS,40.46,0.00,40.46,40.46,inf,-,inf,inf
LLS-GP,39.51,0.32,38.94,39.65,5.79,0.01,5.77,5.80
LLS-GP-SepKer,40.46,0.00,40.46,40.46,5.24,0.07,5.12,5.29
it-LLS-GP,39.86,0.76,39.02,40.42,5.57,0.03,5.55,5.60
LLS-BNN,55.11,0.96,53.91,56.50,inf,-,inf,inf
it-LLS-BNN,52.37,2.84,49.03,56.49,inf,-,inf,inf
SVR,40.99,0.00,40.99,40.99,inf,-,inf,inf
SVR-GP,40.99,0.00,40.99,40.99,5.58,0.00,5.58,5.58
SVR-GP-SepKer,40.99,0.00,40.99,40.99,5.20,0.07,5.16,5.32
it-SVR-GP,41.74,1.03,40.99,42.86,5.59,0.01,5.58,5.60
SVR-BNN,50.49,1.74,48.68,53.09,inf,-,inf,inf
it-SVR-BNN,51.10,3.04,48.61,55.58,inf,-,inf,inf
GP,37.39,3.60,30.95,39.00,5.74,0.33,5.14,5.88
GP-SepKer,41.23,0.23,40.94,41.58,5.17,0.04,5.14,5.23
BNN,41.96,0.23,41.57,42.20,6.94,0.32,6.60,7.37
SPGP,28.53,0.01,28.52,28.55,5.17,0.00,5.17,5.17
SPGP-SepKer,28.58,0.00,28.58,28.58,4.92,0.00,4.92,4.92
SPGP-from-ones,41.57,0.00,41.57,41.57,6.16,0.00,6.16,6.16
BaMbANN,40.51,0.15,40.28,40.70,inf,-,inf,inf
\end{filecontents*}
\csvreader[respect underscore=true, head to column names]{./csvs/simdyn_stats_exp_interp_SENSING_dim_0.csv}{}
{\\\algo & \rmsemean ~$\pm$ \rmsestd & \rmseamin ~/ \rmseamax & \nllhmean ~$\pm$ \nllhstd & \nllhamin ~/ \nllhamax}
\end{tabular}
\end{table*}

\begin{table*}
\centering
\caption {\label{tab:simdynglobalinterone} Results \simdynglobalinter{}, joint 1. Statistics are built over five independent executions of each algorithm.} 
\begin{tabular}{l||ll|ll}
Method & \multicolumn{4}{c}{\simdynglobalinter{}, joint 1} \\
\hline
     & \multicolumn{2}{c|}{RMSE} & \multicolumn{2}{c}{NLLH} \\ 
\cline{2-5}
     &  mean $\pm$ std. & min / max & mean $\pm$ std. & min / max \\
\hline
\begin{filecontents*}{./csvs/simdyn_stats_exp_interp_SENSING_dim_1.csv}
algo,rmsemean,rmsestd,rmseamin,rmseamax,nllhmean,nllhstd,nllhamin,nllhamax
LLS-simulated,14.92,-,14.92,14.92,inf,-,inf,inf
LLS-prior,16.39,-,16.39,16.39,inf,-,inf,inf
LLS,14.88,0.00,14.88,14.88,inf,-,inf,inf
LLS-GP,18.13,0.11,17.94,18.18,4.39,0.01,4.37,4.39
LLS-GP-SepKer,14.88,0.00,14.88,14.88,4.16,0.03,4.12,4.18
it-LLS-GP,15.98,1.52,14.87,17.65,4.32,0.05,4.28,4.37
LLS-BNN,22.70,0.14,22.46,22.81,inf,-,24.66,inf
it-LLS-BNN,23.02,0.23,22.66,23.27,inf,-,27.75,inf
SVR,14.97,0.00,14.97,14.97,inf,-,inf,inf
SVR-GP,14.97,0.00,14.97,14.97,4.28,0.00,4.28,4.28
SVR-GP-SepKer,14.97,0.00,14.97,14.97,4.18,0.05,4.13,4.24
it-SVR-GP,15.21,0.33,14.97,15.57,4.32,0.05,4.28,4.37
SVR-BNN,22.98,0.32,22.58,23.29,inf,-,21.62,inf
it-SVR-BNN,23.47,0.65,22.69,24.15,inf,-,24.35,inf
GP,22.60,3.06,17.12,23.97,4.61,0.19,4.27,4.69
GP-SepKer,15.00,0.10,14.89,15.11,4.23,0.04,4.16,4.29
BNN,15.20,0.17,14.97,15.34,4.83,0.10,4.69,4.94
SPGP,16.36,0.00,16.36,16.37,4.22,0.00,4.22,4.22
SPGP-SepKer,10.35,0.00,10.35,10.35,3.86,0.00,3.86,3.86
SPGP-from-ones,14.36,0.00,14.36,14.36,4.16,0.00,4.16,4.16
BaMbANN,14.76,0.02,14.73,14.79,16.20,1.16,14.88,17.95
\end{filecontents*}
\csvreader[respect underscore=true, head to column names]{./csvs/simdyn_stats_exp_interp_SENSING_dim_1.csv}{}
{\\\algo & \rmsemean ~$\pm$ \rmsestd & \rmseamin ~/ \rmseamax & \nllhmean ~$\pm$ \nllhstd & \nllhamin ~/ \nllhamax}
\end{tabular}
\end{table*}

\begin{table*}
\centering
\caption {\label{tab:simdynglobalintertwo} Results \simdynglobalinter{}, joint 2. Statistics are built over five independent executions of each algorithm.} 
\begin{tabular}{l||ll|ll}
Method & \multicolumn{4}{c}{\simdynglobalinter{}, joint 2} \\
\hline
     & \multicolumn{2}{c|}{RMSE} & \multicolumn{2}{c}{NLLH} \\ 
\cline{2-5}
     &  mean $\pm$ std. & min / max & mean $\pm$ std. & min / max \\
\hline
\begin{filecontents*}{./csvs/simdyn_stats_exp_interp_SENSING_dim_2.csv}
algo,rmsemean,rmsestd,rmseamin,rmseamax,nllhmean,nllhstd,nllhamin,nllhamax
LLS-simulated,13.72,-,13.72,13.72,inf,-,inf,inf
LLS-prior,13.86,-,13.86,13.86,inf,-,inf,inf
LLS,13.61,0.00,13.61,13.61,inf,-,inf,inf
LLS-GP,10.46,0.06,10.43,10.56,4.05,0.00,4.04,4.06
LLS-GP-SepKer,3.24,1.91,1.15,4.73,2.58,0.89,1.56,3.32
it-LLS-GP,12.46,1.67,10.63,13.68,4.20,0.07,4.12,4.25
LLS-BNN,22.20,0.50,21.65,22.91,28.61,3.73,23.37,33.22
it-LLS-BNN,22.72,0.75,21.88,23.68,inf,-,23.74,inf
SVR,13.60,0.00,13.60,13.60,inf,-,inf,inf
SVR-GP,13.60,0.00,13.60,13.60,4.25,0.00,4.25,4.25
SVR-GP-SepKer,1.51,0.41,1.18,2.03,2.28,0.87,1.58,3.37
it-SVR-GP,13.53,0.09,13.44,13.60,4.28,0.04,4.25,4.32
SVR-BNN,22.12,1.26,20.80,23.85,30.70,1.92,28.20,33.12
it-SVR-BNN,22.43,0.66,21.67,23.27,inf,-,25.35,inf
GP,20.63,3.20,14.91,22.06,4.53,0.20,4.17,4.62
GP-SepKer,0.85,0.01,0.85,0.86,1.26,0.00,1.25,1.27
BNN,13.80,0.25,13.66,14.24,9.20,0.54,8.45,9.75
SPGP,15.01,0.00,15.01,15.02,4.14,0.00,4.14,4.14
SPGP-SepKer,13.86,0.00,13.86,13.86,4.05,0.00,4.05,4.05
SPGP-from-ones,13.44,0.00,13.44,13.44,4.13,0.00,4.13,4.13
BaMbANN,13.54,0.01,13.53,13.55,inf,-,74.13,inf
\end{filecontents*}
\csvreader[respect underscore=true, head to column names]{./csvs/simdyn_stats_exp_interp_SENSING_dim_2.csv}{}
{\\\algo & \rmsemean ~$\pm$ \rmsestd & \rmseamin ~/ \rmseamax & \nllhmean ~$\pm$ \nllhstd & \nllhamin ~/ \nllhamax}
\end{tabular}
\end{table*}

\begin{table*}
\centering
\caption {\label{tab:simdynglobalextra} Results \simdynglobalextra{}, joint 0. Statistics are built over five independent executions of each algorithm.} 
\begin{tabular}{l||ll|ll}
Method & \multicolumn{4}{c}{\simdynglobalextra{}, joint 0} \\
\hline
     & \multicolumn{2}{c|}{RMSE} & \multicolumn{2}{c}{NLLH} \\ 
\cline{2-5}
     &  mean $\pm$ std. & min / max & mean $\pm$ std. & min / max \\
\hline
\begin{filecontents*}{./csvs/simdyn_stats_exp_extrap_SENSING_dim_0.csv}
algo,rmsemean,rmsestd,rmseamin,rmseamax,nllhmean,nllhstd,nllhamin,nllhamax
LLS-simulated,35.76,-,35.76,35.76,inf,-,inf,inf
LLS-prior,37.22,-,37.22,37.22,inf,-,inf,inf
LLS,35.27,0.00,35.27,35.27,inf,-,inf,inf
LLS-GP,29.81,0.24,29.37,29.91,4.91,0.01,4.89,4.91
LLS-GP-SepKer,35.27,0.00,35.27,35.27,5.02,0.02,4.99,5.04
it-LLS-GP,33.18,2.95,29.95,35.34,5.08,0.17,4.89,5.20
LLS-BNN,61.44,2.63,57.38,63.70,inf,-,inf,inf
it-LLS-BNN,55.95,3.03,52.28,60.72,inf,-,inf,inf
SVR,35.53,0.00,35.53,35.53,inf,-,inf,inf
SVR-GP,35.53,0.00,35.53,35.53,5.21,0.00,5.21,5.21
SVR-GP-SepKer,35.53,0.00,35.53,35.53,5.01,0.03,4.99,5.05
it-SVR-GP,36.68,1.59,35.52,38.42,5.28,0.09,5.21,5.38
SVR-BNN,55.99,6.99,49.10,65.66,inf,-,inf,inf
it-SVR-BNN,60.72,12.81,44.97,78.88,inf,-,inf,inf
GP,87.85,4.89,79.10,90.04,8.39,0.03,8.33,8.41
GP-SepKer,38.80,0.32,38.42,39.29,5.09,0.03,5.07,5.13
BNN,37.57,0.77,36.48,38.24,5.96,0.17,5.82,6.25
SPGP,27.53,0.01,27.51,27.53,5.02,0.00,5.02,5.02
SPGP-SepKer,27.25,0.00,27.25,27.25,4.80,0.00,4.80,4.80
SPGP-from-ones,36.66,0.00,36.66,36.66,5.65,0.00,5.65,5.65
BaMbANN,35.73,0.21,35.53,36.05,inf,-,inf,inf
\end{filecontents*}
\csvreader[respect underscore=true, head to column names]{./csvs/simdyn_stats_exp_extrap_SENSING_dim_0.csv}{}
{\\\algo & \rmsemean ~$\pm$ \rmsestd & \rmseamin ~/ \rmseamax & \nllhmean ~$\pm$ \nllhstd & \nllhamin ~/ \nllhamax}
\end{tabular}
\end{table*}

\begin{table*}
\centering
\caption {\label{tab:simdynglobalextraone} Results \simdynglobalextra{}, joint 1. Statistics are built over five independent executions of each algorithm.} 
\begin{tabular}{l||ll|ll}
Method & \multicolumn{4}{c}{\simdynglobalextra{}, joint 1} \\
\hline
     & \multicolumn{2}{c|}{RMSE} & \multicolumn{2}{c}{NLLH} \\ 
\cline{2-5}
     &  mean $\pm$ std. & min / max & mean $\pm$ std. & min / max \\
\hline
\begin{filecontents*}{./csvs/simdyn_stats_exp_extrap_SENSING_dim_1.csv}
algo,rmsemean,rmsestd,rmseamin,rmseamax,nllhmean,nllhstd,nllhamin,nllhamax
LLS-simulated,14.94,-,14.94,14.94,inf,-,inf,inf
LLS-prior,16.44,-,16.44,16.44,inf,-,inf,inf
LLS,14.90,0.00,14.90,14.90,inf,-,inf,inf
LLS-GP,16.63,0.06,16.52,16.66,4.36,0.01,4.34,4.37
LLS-GP-SepKer,14.90,0.00,14.90,14.90,4.16,0.03,4.12,4.18
it-LLS-GP,15.49,0.82,14.89,16.39,4.31,0.05,4.28,4.37
LLS-BNN,26.22,3.23,23.68,31.70,inf,-,27.32,inf
it-LLS-BNN,28.37,2.05,25.86,30.85,inf,-,31.43,inf
SVR,15.02,0.00,15.02,15.02,inf,-,inf,inf
SVR-GP,15.02,0.00,15.02,15.02,4.28,0.00,4.28,4.28
SVR-GP-SepKer,15.02,0.00,15.02,15.02,4.18,0.05,4.13,4.24
it-SVR-GP,15.17,0.21,15.02,15.40,4.33,0.06,4.28,4.39
SVR-BNN,27.88,3.18,24.52,31.46,inf,-,25.67,inf
it-SVR-BNN,32.06,0.58,31.16,32.58,inf,-,26.69,inf
GP,40.74,1.65,37.80,41.48,5.09,0.04,5.02,5.11
GP-SepKer,15.34,0.14,15.18,15.47,4.24,0.04,4.18,4.30
BNN,15.33,0.15,15.11,15.54,4.87,0.13,4.69,5.06
SPGP,16.05,0.00,16.05,16.05,4.21,0.00,4.21,4.21
SPGP-SepKer,12.26,0.00,12.26,12.26,3.96,0.00,3.96,3.96
SPGP-from-ones,14.38,0.00,14.38,14.38,4.16,0.00,4.16,4.16
BaMbANN,14.79,0.02,14.77,14.81,16.47,2.07,13.86,19.58
\end{filecontents*}
\csvreader[respect underscore=true, head to column names]{./csvs/simdyn_stats_exp_extrap_SENSING_dim_1.csv}{}
{\\\algo & \rmsemean ~$\pm$ \rmsestd & \rmseamin ~/ \rmseamax & \nllhmean ~$\pm$ \nllhstd & \nllhamin ~/ \nllhamax}
\end{tabular}
\end{table*}

\begin{table*}
\centering
\caption {\label{tab:simdynglobalextratwo} Results \simdynglobalextra{}, joint 2. Statistics are built over five independent executions of each algorithm.} 
\begin{tabular}{l||ll|ll}
Method & \multicolumn{4}{c}{\simdynglobalextra{}, joint 2} \\
\hline
     & \multicolumn{2}{c|}{RMSE} & \multicolumn{2}{c}{NLLH} \\ 
\cline{2-5}
     &  mean $\pm$ std. & min / max & mean $\pm$ std. & min / max \\
\hline
\begin{filecontents*}{./csvs/simdyn_stats_exp_extrap_SENSING_dim_2.csv}
algo,rmsemean,rmsestd,rmseamin,rmseamax,nllhmean,nllhstd,nllhamin,nllhamax
LLS-simulated,13.72,-,13.72,13.72,inf,-,inf,inf
LLS-prior,13.85,-,13.85,13.85,inf,-,inf,inf
LLS,13.60,0.00,13.60,13.60,inf,-,inf,inf
LLS-GP,11.55,0.04,11.53,11.61,4.18,0.01,4.16,4.18
LLS-GP-SepKer,3.25,1.91,1.16,4.73,2.58,0.89,1.57,3.32
it-LLS-GP,12.88,1.09,11.69,13.68,4.24,0.01,4.22,4.25
LLS-BNN,23.91,1.10,22.76,25.47,inf,-,25.88,inf
it-LLS-BNN,24.44,1.21,23.11,25.51,inf,-,24.99,inf
SVR,13.60,0.00,13.60,13.60,inf,-,inf,inf
SVR-GP,13.60,0.00,13.60,13.60,4.25,0.00,4.25,4.25
SVR-GP-SepKer,1.52,0.41,1.18,2.04,2.28,0.87,1.58,3.37
it-SVR-GP,13.57,0.04,13.53,13.60,4.29,0.06,4.25,4.35
SVR-BNN,24.95,2.68,22.78,28.03,inf,-,27.28,inf
it-SVR-BNN,25.12,1.72,23.41,27.73,inf,-,28.27,inf
GP,24.08,3.47,17.88,25.63,4.70,0.21,4.33,4.79
GP-SepKer,0.86,0.00,0.86,0.87,1.27,0.00,1.26,1.28
BNN,13.91,0.29,13.73,14.42,9.03,0.61,8.23,9.83
SPGP,14.61,0.01,14.60,14.62,4.14,0.00,4.14,4.14
SPGP-SepKer,13.70,0.00,13.70,13.70,4.04,0.00,4.04,4.04
SPGP-from-ones,13.42,0.00,13.42,13.42,4.13,0.00,4.13,4.13
BaMbANN,13.56,0.01,13.54,13.57,inf,-,inf,inf
\end{filecontents*}
\csvreader[respect underscore=true, head to column names]{./csvs/simdyn_stats_exp_extrap_SENSING_dim_2.csv}{}
{\\\algo & \rmsemean ~$\pm$ \rmsestd & \rmseamin ~/ \rmseamax & \nllhmean ~$\pm$ \nllhstd & \nllhamin ~/ \nllhamax}
\end{tabular}
\end{table*}

\begin{table*}
\centering
\caption {\label{tab:simdynlocalinter} Results \simdynlocalinter{}, joint 0. Statistics are built over five independent executions of each algorithm.} 
\begin{tabular}{l||ll|ll}
Method & \multicolumn{4}{c}{\simdynlocalinter{}, joint 0} \\
\hline
     & \multicolumn{2}{c|}{RMSE} & \multicolumn{2}{c}{NLLH} \\ 
\cline{2-5}
     &  mean $\pm$ std. & min / max & mean $\pm$ std. & min / max \\
\hline
\begin{filecontents*}{./csvs/simdyn_stats_exp_interp_SENSING_dist_local_dim_0.csv}
algo,rmsemean,rmsestd,rmseamin,rmseamax,nllhmean,nllhstd,nllhamin,nllhamax
LLS-simulated,39.42,-,39.42,39.42,inf,-,inf,inf
LLS-prior,40.33,-,40.33,40.33,inf,-,inf,inf
LLS,38.43,0.00,38.43,38.43,inf,-,inf,inf
LLS-GP,28.82,0.00,28.82,28.82,inf,-,inf,inf
LLS-GP-SepKer,38.43,0.00,38.43,38.43,5.09,0.00,5.09,5.10
it-LLS-GP,29.02,0.00,29.02,29.02,inf,-,inf,inf
LLS-BNN,44.48,1.74,41.60,45.73,inf,-,inf,inf
it-LLS-BNN,45.67,2.18,43.35,47.95,inf,-,inf,inf
SVR,39.05,0.00,39.05,39.05,inf,-,inf,inf
SVR-GP,39.07,0.01,39.06,39.08,inf,-,inf,inf
SVR-GP-SepKer,39.05,0.00,39.05,39.05,5.11,0.01,5.09,5.12
it-SVR-GP,39.07,0.00,39.07,39.07,inf,-,inf,inf
SVR-BNN,43.06,1.04,42.13,44.84,inf,-,inf,inf
it-SVR-BNN,42.83,1.49,40.91,44.26,inf,-,inf,inf
GP,53.52,10.37,45.95,64.88,inf,-,inf,inf
GP-SepKer,38.87,0.03,38.84,38.91,5.08,0.00,5.08,5.09
BNN,38.91,0.15,38.69,39.11,7.03,0.35,6.54,7.37
SPGP,19.07,0.00,19.06,19.07,4.67,0.00,4.67,4.67
SPGP-SepKer,22.54,0.00,22.54,22.54,4.62,0.00,4.62,4.62
SPGP-from-ones,22.87,0.00,22.87,22.87,5.39,0.00,5.39,5.39
BaMbANN,37.26,1.13,36.30,38.64,inf,-,7.31,inf
\end{filecontents*}
\csvreader[respect underscore=true, head to column names]{./csvs/simdyn_stats_exp_interp_SENSING_dist_local_dim_0.csv}{}
{\\\algo & \rmsemean ~$\pm$ \rmsestd & \rmseamin ~/ \rmseamax & \nllhmean ~$\pm$ \nllhstd & \nllhamin ~/ \nllhamax}
\end{tabular}
\end{table*}

\begin{table*}
\centering
\caption {\label{tab:simdynlocalinterone} Results \simdynlocalinter{}, joint 1. Statistics are built over five independent executions of each algorithm.} 
\begin{tabular}{l||ll|ll}
Method & \multicolumn{4}{c}{\simdynlocalinter{}, joint 1} \\
\hline
     & \multicolumn{2}{c|}{RMSE} & \multicolumn{2}{c}{NLLH} \\ 
\cline{2-5}
     &  mean $\pm$ std. & min / max & mean $\pm$ std. & min / max \\
\hline
\begin{filecontents*}{./csvs/simdyn_stats_exp_interp_SENSING_dist_local_dim_1.csv}
algo,rmsemean,rmsestd,rmseamin,rmseamax,nllhmean,nllhstd,nllhamin,nllhamax
LLS-simulated,6.10,-,6.10,6.10,inf,-,inf,inf
LLS-prior,9.50,-,9.50,9.50,inf,-,inf,inf
LLS,6.10,0.00,6.10,6.10,inf,-,inf,inf
LLS-GP,5.38,0.00,5.38,5.38,3.17,0.00,3.17,3.17
LLS-GP-SepKer,5.50,0.75,4.51,6.04,3.31,0.30,2.95,3.53
it-LLS-GP,5.43,0.00,5.43,5.43,3.17,0.00,3.17,3.17
LLS-BNN,7.69,0.18,7.47,7.94,inf,-,18.86,inf
it-LLS-BNN,7.60,0.36,7.22,8.14,inf,-,18.46,inf
SVR,6.05,0.00,6.05,6.05,inf,-,inf,inf
SVR-GP,6.06,0.00,6.06,6.06,3.93,0.00,3.93,3.93
SVR-GP-SepKer,5.88,0.26,5.42,6.03,3.19,0.05,3.11,3.22
it-SVR-GP,6.06,0.00,6.06,6.06,3.93,0.00,3.93,3.93
SVR-BNN,7.44,0.25,7.14,7.82,inf,-,17.52,inf
it-SVR-BNN,7.79,0.39,7.36,8.31,inf,-,19.52,inf
GP,23.79,6.44,19.09,30.84,4.46,0.33,4.22,4.83
GP-SepKer,5.46,0.04,5.41,5.50,3.09,0.01,3.08,3.11
BNN,6.26,0.12,6.10,6.40,3.30,0.02,3.27,3.31
SPGP,5.51,0.00,5.51,5.52,3.48,0.00,3.48,3.48
SPGP-SepKer,4.58,0.00,4.58,4.58,2.98,0.00,2.98,2.98
SPGP-from-ones,5.21,0.00,5.21,5.21,3.41,0.00,3.41,3.41
BaMbANN,6.06,0.05,6.02,6.16,6.53,3.12,3.80,11.61
\end{filecontents*}
\csvreader[respect underscore=true, head to column names]{./csvs/simdyn_stats_exp_interp_SENSING_dist_local_dim_1.csv}{}
{\\\algo & \rmsemean ~$\pm$ \rmsestd & \rmseamin ~/ \rmseamax & \nllhmean ~$\pm$ \nllhstd & \nllhamin ~/ \nllhamax}
\end{tabular}
\end{table*}

\begin{table*}
\centering
\caption {\label{tab:simdynlocalintertwo} Results \simdynlocalinter{}, joint 2. Statistics are built over five independent executions of each algorithm.} 
\begin{tabular}{l||ll|ll}
Method & \multicolumn{4}{c}{\simdynlocalinter{}, joint 2} \\
\hline
     & \multicolumn{2}{c|}{RMSE} & \multicolumn{2}{c}{NLLH} \\ 
\cline{2-5}
     &  mean $\pm$ std. & min / max & mean $\pm$ std. & min / max \\
\hline
\begin{filecontents*}{./csvs/simdyn_stats_exp_interp_SENSING_dist_local_dim_2.csv}
algo,rmsemean,rmsestd,rmseamin,rmseamax,nllhmean,nllhstd,nllhamin,nllhamax
LLS-simulated,0.79,-,0.79,0.79,inf,-,inf,inf
LLS-prior,1.30,-,1.30,1.30,inf,-,inf,inf
LLS,0.91,0.00,0.91,0.91,inf,-,inf,inf
LLS-GP,1.07,0.00,1.07,1.07,2.92,0.00,2.92,2.92
LLS-GP-SepKer,0.91,0.00,0.91,0.91,1.33,0.00,1.33,1.33
it-LLS-GP,1.08,0.00,1.08,1.08,2.92,0.00,2.92,2.92
LLS-BNN,1.62,0.10,1.47,1.74,inf,-,21.84,inf
it-LLS-BNN,1.61,0.06,1.54,1.69,inf,-,22.83,inf
SVR,0.78,0.00,0.78,0.78,inf,-,inf,inf
SVR-GP,0.78,0.00,0.78,0.78,3.88,0.00,3.88,3.88
SVR-GP-SepKer,0.78,0.00,0.78,0.78,1.17,0.00,1.17,1.17
it-SVR-GP,0.80,0.01,0.79,0.81,3.88,0.00,3.88,3.88
SVR-BNN,1.58,0.07,1.50,1.65,inf,-,24.93,inf
it-SVR-BNN,1.55,0.06,1.49,1.63,inf,-,25.35,inf
GP,3.00,0.23,2.84,3.25,3.38,0.15,3.27,3.55
GP-SepKer,0.76,0.01,0.75,0.77,1.15,0.01,1.13,1.16
BNN,1.02,0.03,0.96,1.05,1.84,0.03,1.81,1.88
SPGP,0.92,0.00,0.92,0.92,3.37,0.00,3.37,3.37
SPGP-SepKer,0.76,0.00,0.76,0.76,1.15,0.00,1.15,1.15
SPGP-from-ones,1.22,0.00,1.22,1.22,3.30,0.00,3.30,3.30
BaMbANN,0.80,0.01,0.79,0.82,1.31,0.07,1.23,1.41
\end{filecontents*}
\csvreader[respect underscore=true, head to column names]{./csvs/simdyn_stats_exp_interp_SENSING_dist_local_dim_2.csv}{}
{\\\algo & \rmsemean ~$\pm$ \rmsestd & \rmseamin ~/ \rmseamax & \nllhmean ~$\pm$ \nllhstd & \nllhamin ~/ \nllhamax}
\end{tabular}
\end{table*}

\begin{table*}
\centering
\caption {\label{tab:simdynlocalextra} Results \simdynlocalextra{}, joint 0. Statistics are built over five independent executions of each algorithm.} 
\begin{tabular}{l||ll|ll}
Method & \multicolumn{4}{c}{\simdynlocalextra{}, joint 0} \\
\hline
     & \multicolumn{2}{c|}{RMSE} & \multicolumn{2}{c}{NLLH} \\ 
\cline{2-5}
     &  mean $\pm$ std. & min / max & mean $\pm$ std. & min / max \\
\hline
\begin{filecontents*}{./csvs/simdyn_stats_exp_extrap_SENSING_dist_local_dim_0.csv}
algo,rmsemean,rmsestd,rmseamin,rmseamax,nllhmean,nllhstd,nllhamin,nllhamax
LLS-simulated,32.95,-,32.95,32.95,inf,-,inf,inf
LLS-prior,34.58,-,34.58,34.58,inf,-,inf,inf
LLS,32.36,0.00,32.36,32.36,inf,-,inf,inf
LLS-GP,31.22,0.00,31.22,31.22,6.22,0.00,6.22,6.22
LLS-GP-SepKer,32.36,0.00,32.36,32.36,4.90,0.00,4.90,4.90
it-LLS-GP,31.32,0.00,31.32,31.32,6.24,0.00,6.24,6.24
LLS-BNN,41.66,7.49,34.89,53.34,inf,-,inf,inf
it-LLS-BNN,41.42,4.06,35.85,45.66,inf,-,inf,inf
SVR,32.72,0.00,32.72,32.72,inf,-,inf,inf
SVR-GP,32.69,0.01,32.69,32.70,5.31,0.00,5.31,5.31
SVR-GP-SepKer,32.72,0.00,32.72,32.72,4.91,0.01,4.91,4.92
it-SVR-GP,32.68,0.00,32.68,32.68,5.31,0.00,5.31,5.31
SVR-BNN,37.45,3.24,33.00,40.91,inf,-,inf,inf
it-SVR-BNN,42.94,5.38,36.46,50.22,inf,-,inf,inf
GP,112.80,4.46,109.54,117.69,10.81,1.92,8.71,12.21
GP-SepKer,35.94,0.06,35.86,36.03,5.00,0.00,5.00,5.00
BNN,33.00,0.58,32.24,33.66,5.81,0.10,5.66,5.89
SPGP,24.76,0.00,24.76,24.77,5.15,0.00,5.15,5.15
SPGP-SepKer,31.81,0.00,31.81,31.81,5.25,0.00,5.25,5.25
SPGP-from-ones,24.00,0.00,23.99,24.00,5.35,0.00,5.35,5.35
BaMbANN,32.64,7.52,25.02,43.06,inf,-,5.33,inf
\end{filecontents*}
\csvreader[respect underscore=true, head to column names]{./csvs/simdyn_stats_exp_extrap_SENSING_dist_local_dim_0.csv}{}
{\\\algo & \rmsemean ~$\pm$ \rmsestd & \rmseamin ~/ \rmseamax & \nllhmean ~$\pm$ \nllhstd & \nllhamin ~/ \nllhamax}
\end{tabular}
\end{table*}

\begin{table*}
\centering
\caption {\label{tab:simdynlocalextraone} Results \simdynlocalextra{}, joint 1. Statistics are built over five independent executions of each algorithm.} 
\begin{tabular}{l||ll|ll}
Method & \multicolumn{4}{c}{\simdynlocalextra{}, joint 1} \\
\hline
     & \multicolumn{2}{c|}{RMSE} & \multicolumn{2}{c}{NLLH} \\ 
\cline{2-5}
     &  mean $\pm$ std. & min / max & mean $\pm$ std. & min / max \\
\hline
\begin{filecontents*}{./csvs/simdyn_stats_exp_extrap_SENSING_dist_local_dim_1.csv}
algo,rmsemean,rmsestd,rmseamin,rmseamax,nllhmean,nllhstd,nllhamin,nllhamax
LLS-simulated,6.09,-,6.09,6.09,inf,-,inf,inf
LLS-prior,9.54,-,9.54,9.54,inf,-,inf,inf
LLS,6.09,0.00,6.09,6.09,inf,-,inf,inf
LLS-GP,9.21,0.00,9.21,9.21,4.14,0.00,4.14,4.14
LLS-GP-SepKer,6.44,1.09,5.73,8.37,3.50,0.16,3.23,3.67
it-LLS-GP,9.25,0.00,9.25,9.25,4.14,0.00,4.14,4.14
LLS-BNN,12.56,2.26,9.87,15.19,inf,-,40.43,inf
it-LLS-BNN,13.15,2.29,10.37,16.29,inf,-,23.25,inf
SVR,6.04,0.00,6.04,6.04,inf,-,inf,inf
SVR-GP,6.04,0.00,6.04,6.05,3.93,0.00,3.93,3.93
SVR-GP-SepKer,5.87,0.28,5.38,6.03,3.19,0.05,3.10,3.22
it-SVR-GP,6.05,0.00,6.05,6.05,3.93,0.00,3.93,3.93
SVR-BNN,12.17,1.04,11.11,13.82,inf,-,inf,inf
it-SVR-BNN,13.58,1.53,11.98,15.31,inf,-,32.21,inf
GP,49.31,3.18,46.99,52.80,5.41,0.03,5.39,5.44
GP-SepKer,12.36,2.79,10.03,17.20,4.59,0.62,4.06,5.64
BNN,6.45,0.18,6.26,6.68,3.33,0.03,3.30,3.36
SPGP,7.17,0.01,7.16,7.19,3.68,0.00,3.67,3.68
SPGP-SepKer,6.67,0.00,6.67,6.67,3.32,0.00,3.32,3.32
SPGP-from-ones,6.35,0.00,6.35,6.35,3.56,0.00,3.56,3.56
BaMbANN,6.08,0.05,6.03,6.13,5.59,2.14,3.78,9.07
\end{filecontents*}
\csvreader[respect underscore=true, head to column names]{./csvs/simdyn_stats_exp_extrap_SENSING_dist_local_dim_1.csv}{}
{\\\algo & \rmsemean ~$\pm$ \rmsestd & \rmseamin ~/ \rmseamax & \nllhmean ~$\pm$ \nllhstd & \nllhamin ~/ \nllhamax}
\end{tabular}
\end{table*}

\begin{table*}
\centering
\caption {\label{tab:simdynlocalextratwo} Results \simdynlocalextra{}, joint 2. Statistics are built over five independent executions of each algorithm.} 
\begin{tabular}{l||ll|ll}
Method & \multicolumn{4}{c}{\simdynlocalextra{}, joint 2} \\
\hline
     & \multicolumn{2}{c|}{RMSE} & \multicolumn{2}{c}{NLLH} \\ 
\cline{2-5}
     &  mean $\pm$ std. & min / max & mean $\pm$ std. & min / max \\
\hline
\begin{filecontents*}{./csvs/simdyn_stats_exp_extrap_SENSING_dist_local_dim_2.csv}
algo,rmsemean,rmsestd,rmseamin,rmseamax,nllhmean,nllhstd,nllhamin,nllhamax
LLS-simulated,0.78,-,0.78,0.78,inf,-,inf,inf
LLS-prior,1.29,-,1.29,1.29,inf,-,inf,inf
LLS,0.88,0.00,0.88,0.88,inf,-,inf,inf
LLS-GP,2.06,0.00,2.06,2.06,3.96,0.00,3.96,3.96
LLS-GP-SepKer,0.88,0.00,0.88,0.88,1.29,0.00,1.29,1.30
it-LLS-GP,2.07,0.00,2.07,2.07,3.96,0.00,3.96,3.96
LLS-BNN,1.87,0.21,1.68,2.13,inf,-,21.28,inf
it-LLS-BNN,1.83,0.20,1.65,2.14,27.38,4.27,20.41,31.16
SVR,0.78,0.00,0.78,0.78,inf,-,inf,inf
SVR-GP,0.78,0.00,0.78,0.78,3.88,0.00,3.88,3.88
SVR-GP-SepKer,0.78,0.00,0.78,0.78,1.17,0.00,1.17,1.17
it-SVR-GP,0.80,0.01,0.78,0.81,3.88,0.00,3.88,3.88
SVR-BNN,1.87,0.06,1.80,1.93,inf,-,22.90,inf
it-SVR-BNN,1.81,0.13,1.67,2.00,inf,-,25.19,inf
GP,12.59,0.44,12.27,13.07,4.37,0.29,4.16,4.69
GP-SepKer,0.76,0.01,0.75,0.77,1.15,0.01,1.13,1.16
BNN,1.05,0.06,0.97,1.14,1.88,0.03,1.83,1.91
SPGP,1.07,0.01,1.05,1.07,3.54,0.00,3.54,3.54
SPGP-SepKer,0.76,0.00,0.76,0.76,1.15,0.00,1.15,1.15
SPGP-from-ones,2.13,0.00,2.13,2.13,3.44,0.00,3.44,3.44
BaMbANN,0.83,0.06,0.79,0.92,1.36,0.07,1.27,1.43
\end{filecontents*}
\csvreader[respect underscore=true, head to column names]{./csvs/simdyn_stats_exp_extrap_SENSING_dist_local_dim_2.csv}{}
{\\\algo & \rmsemean ~$\pm$ \rmsestd & \rmseamin ~/ \rmseamax & \nllhmean ~$\pm$ \nllhstd & \nllhamin ~/ \nllhamax}
\end{tabular}
\end{table*}